\documentclass[pdflatex,sn-mathphys]{sn-jnl}

\usepackage{longtable}
\usepackage{mathtools}
\usepackage{enumitem}
\usepackage{commath}
\usepackage{graphicx}
\usepackage{amsmath}

\jyear{2022}
\theoremstyle{thmstyleone}

\theoremstyle{thmstyletwo}

\theoremstyle{thmstylethree}

\usepackage{tabularx}

\usepackage{booktabs, multirow} 
\usepackage{soul}
\usepackage{textcomp} 
\usepackage{amssymb} 

\begin{document}
\title[Article Title]{Unsupervised learning of Data-driven Facial Expression Coding System (DFECS) using keypoint tracking}

\author*[1]{\fnm{Shivansh Chandra} \sur{Tripathi}}\email{shivansh.tripathi.iitd@@gmail.com}

\author[1,2,3]{\fnm{Rahul} \sur{Garg}}

\affil[1]{\orgdiv{Department of Computer Science and Engineering}, \orgname{Indian Institute of Technology Delhi}, \city{Delhi}, \postcode{110016}, \country{India}}

\affil[2]{\orgdiv{Amar Nath and Shashi Khosla School of Information Technology}, \orgname{Indian Institute of Technology Delhi}, \city{Delhi}, \postcode{110016}, \country{India}}

\affil[3]{\orgdiv{National Resource Centre for Value Education in Engineering}, \orgname{Indian Institute of Technology Delhi}, \city{Delhi}, \postcode{110016}, \country{India}}

\abstract{The development of existing facial coding systems, such as the Facial Action Coding System (FACS), relied on manual examination of facial expression videos for defining Action Units (AUs). To overcome the labor-intensive nature of this process, we propose the unsupervised learning of an automated facial coding system by leveraging computer-vision-based facial keypoint tracking. In this novel facial coding system called the Data-driven Facial Expression Coding System (DFECS), the AUs are estimated by applying dimensionality reduction to facial keypoint movements from a neutral frame through a proposed Full Face Model (FFM). FFM employs a two-level decomposition using advanced dimensionality reduction techniques such as dictionary learning (DL) and non-negative matrix factorization (NMF). These techniques enhance the interpretability of AUs by introducing constraints such as sparsity and positivity to the encoding matrix. Results show that DFECS AUs estimated from the DISFA dataset can account for an average variance of up to 91.29 percent in test datasets (CK+ and BP4D-Spontaneous) and also surpass the variance explained by keypoint-based equivalents of FACS AUs in these datasets. Additionally, 87.5 percent of DFECS AUs are interpretable, i.e., align with the direction of facial muscle movements. In summary, advancements in automated facial coding systems can accelerate facial expression analysis across diverse fields such as security, healthcare, and entertainment. These advancements offer numerous benefits, including enhanced detection of abnormal behavior, improved pain analysis in healthcare settings, and enriched emotion-driven interactions. To facilitate further research, the code repository of DFECS has been made publicly accessible.}

\keywords{Facial Expressions, Action Units (AUs), Facial Action Coding System (FACS), Facial Keypoints, PCA AUs, DFECS AUs}

\maketitle

\section{Introduction}
\label{intro}

Interpersonal communication encompasses both verbal (speech) and non-verbal modes, such as facial expressions and body language.\footnote{Code available here: \url{https://github.com/Shivansh-ct/DFECS-AUs}} Among these non-verbal cues, facial expressions play a pivotal role in social interactions and emotional intelligence, serving as a window into the human mind. Examples of commonly observed facial expressions include smiles, frowns, wrinkled noses, and raised eyebrows. Additionally, facial expressions can reveal the genuine emotional state of a person, making them valuable for applications such as lie detection~\cite{porter2008reading,ekman2009lie} and the examination of medical conditions like depression~\cite{huang2021elderly}. Moreover, they are essential for effectively conveying and understanding emotions in complex social interactions~\cite{zhi2020comprehensive}. The study of facial expressions extends across diverse disciplines, including psychology, neuroscience, sociology, human-computer interaction, healthcare, and affective computing~\cite{ekman1978facial, ekman2002facial, rosenberg2020face, tripathi2024consistent, zhi2020comprehensive}.

\par Research indicates that our ability to interpret emotions from facial expressions relies on specific movements of facial muscles~\cite{wehrle2000studying,matsumoto2008facial}. These subtle variations in facial muscle movements play a crucial role in various studies on prototypical emotions such as happiness, sadness, surprise, anger, and more~\cite{ekman2002facial}. Additionally, they are instrumental in distinguishing between genuine and fake smiles, characterizing conditions like pain, depression, stroke, Parkinson's, autism, and schizophrenia~\cite{hamm2011automated,ekman1988smiles,hess1994cues,zhi2020comprehensive,heilman1993emotional,blonder2005affective,schimmel2011quantitative,tripathi2023protocol,jin2020diagnosing,trevisan2018facial,kohler2010facial,williams2016updating}. For example, the expression of happiness often involves the muscles pulling the lip corners diagonally upwards and creating wrinkles around the eyes, while disgust is characterized by muscles causing nose wrinkling~\cite{ekman2002facial}. A genuine smile includes muscles that raise the cheeks towards the eyelids, producing wrinkles around the eyes known as crow's feet, which is often absent in fake smiles~\cite{ekman1988smiles,hess1994cues}. Furthermore, paralyzed or drooping facial muscles can serve as indicators of neurological disorders such as stroke~\cite{heilman1993emotional,blonder2005affective,schimmel2011quantitative,tripathi2023protocol}. In conditions like Parkinson's disease, there is a significant reduction in facial expression movements, and local tremors in facial muscles may occur~\cite{jin2020diagnosing}. Variations in facial expressions between individuals with autism spectrum disorder and healthy subjects have been documented~\cite{trevisan2018facial}. Subjects with schizophrenia exhibit abnormal expressions and also report impaired perception of facial expressions of emotion~\cite{hamm2011automated,kohler2010facial}. Facial muscle movements also play a role in characterizing pain, aiding in pain detection, especially when patients are unable to communicate verbally~\cite{chen2018automated}. Therefore, the analysis of subtle differences in facial muscle movements holds the potential for diverse applications in facial expression analysis.

\par Psychologists have developed facial coding systems to objectively describe facial muscle movements consistently across various studies on facial expressions~\cite{zhi2020comprehensive}. These systems define a set of atomic expressions which we refer to as the Action Units (AUs), that encode the observable movements of facial muscles rather than delving into the underlying meaning of the displayed expression~\cite{zhi2020comprehensive}. For instance, a specific AU might represent the upward movement of the eyebrows or a particular motion of the lips, such as pulling them diagonally upward in a smile. These AUs can occur independently or in combinations, and a collection of AUs within a facial coding system can effectively capture a broad spectrum of facial expressions.

\par Existing facial coding systems include Facial Affect Scoring Technique (FAST)~\cite{ekman1971facial}, Maximally Discriminative Facial Movement Coding System (MAX)~\cite{izard1979maximally}, Monadic Phases Coding System (MP)~\cite{izard1979maximally, matias1989comparison}, Face Animation Parameters (FAP)~\cite{ostermann2002face}, Facial Expression Coding System (FACES)~\cite{kring2007facial}, Facial Action Coding System (FACS)~\cite{ekman1978facial,ekman2002facial}, Neonatal Facial Coding System (NFCS)~\cite{peters2003neonatal}, and Child Facial Coding System (CFACS)~\cite{gilbert1999postoperative}. These encoding systems have been designed for a variety of applications, for instance, the coding system FAST~\cite{ekman1971facial} is dedicated to encode facial expressions representing six distinct emotions: happiness, anger, surprise, sadness, fear, and disgust. MAX~\cite{izard1979maximally} and MP~\cite{izard1979maximally, matias1989comparison} are designed to encode facial expressions related to infants' affective behavior. FAP~\cite{ostermann2002face} is integrated into the MPEG-4 Face and Body Animation (FBA) standard, serving the purpose of generating facial expressions on virtual faces. FACES~\cite{kring2007facial}, on the other hand, focuses on valence, measuring the intensity of positive or negative emotions conveyed by a facial expression. FACS~\cite{ekman1978facial,ekman2002facial} stands out as a comprehensive coding system capable of encoding various facial expressions, not limited to emotional expressions alone. Extensions of FACS, such as NFCS~\cite{peters2003neonatal} and CFACS~\cite{gilbert1999postoperative}, find applications in pain studies involving infants. Among all these systems, FACS is recognized as the most comprehensive coding system for facial expressions to date~\cite{zhi2020comprehensive,ekman1978facial,ekman2002facial}.

\par All of the existing coding systems, including FAST, MAX, MP, FAP, FACES, FACS, NFCS, and CFACS, were developed by manually analyzing facial expression videos. For instance, Ekman and Friesen dedicated years to closely examining recorded facial muscle movements in order to define the FACS AUs~\cite{ekman1976measuring}. Given the labor-intensive nature of this manual approach, the creation of a comprehensive coding system requires a significant investment of time and effort~\cite{ekman1976measuring}. Moreover, among the manually devised systems, even the most comprehensive one, FACS, covers the majority of visible expressions but not all of them~\cite{ekman1976measuring,ekman1978facial,ekman2002facial}, suggesting the potential for discovering a facial coding system that is even more comprehensive.

\par With the advent of deep learning-based technologies for computer vision, a substantial amount of standardized facial expression videos are readily available. The face images in these videos have been annotated with (x, y) coordinates of selected keypoints representing prominent facial features such as eyes, eyebrows, lips, mouth, and jawline. The collection of these keypoints coordinates within a given face image is illustrated in Fig.~\ref{fig:datasetskp}. The annotation of these keypoints can be achieved automatically using deep learning-based computer vision algorithms~\cite{trigeorgis2016mnemonic,peng2016recurrent,guo2018stabilizing,sun2019fab,liu2017two,nagae2020iterative,wu2021design,dong2018supervision,dong2020supervision}.

The keypoint-labeled facial expression videos represented using keypoint motion vectors (KPMs) (see section~\ref{sec:kpm-generation}) can be encoded with a small number of basis vectors that are biologically interpretable. To our knowledge, the work of Tripathi et al.~\cite{chandra2023pca} marks the first attempt to explore an unsupervised learning approach for deriving a low dimensional representation of KPMs using principal component analysis (PCA) on keypoint labeled videos. Using the principal components of the KPMs as novel Action Units, they proposed the discovery of a new facial coding system called PCA AUs.

The objectives of PCA AUs in their coding system were twofold. Firstly, it should represent any facial keypoint movements with high accuracy. Secondly, the system should maintain biological interpretability. The results of their study indicate that PCA AUs, derived from the keypoint-labeled DISFA dataset~\cite{mavadati2012automatic,mavadati2013disfa}, can effectively encode various facial expression samples across different datasets (CK+~\cite{kanade2000comprehensive,lucey2010extended} and BP4D-Spontaneous~\cite{zhang2013high,zhang2014bp4d}) with an average reconstruction error of 7.17 percent. Furthermore, the coding ability demonstrated by PCA AUs is comparable to that of FACS AUs.

\par However, a notable drawback of PCA AUs is that half of the PCA AUs are non-interpretable. Additionally, the encoding matrix, which contains the linear weights for the linear combinations of PCA AUs to represent a given sample, includes both positive and negative values. This suggests that each PCA AU can exhibit movement in both positive and negative directions from a neutral position. This may not be biologically plausible if AUs are intended to comply with facial muscles that typically move unidirectionally from a neutral face. Furthermore, the dense nature of the encoding matrix implies that all PCA AUs are employed to represent the spatial movement of 68 facial keypoints in any given sample. This is unlikely to be biologically plausible because sometimes only a few of the facial parts move, suggesting that only a few AUs are expected to be present in specific cases.

\par In this work, we aim to enhance the interpretability of AUs in unsupervised automated coding systems by introducing sparsity and positivity constraints to the encoding matrix. To achieve this, we propose a Full Face Model (FFM) that employs a two-level decomposition using advanced dimensionality reduction techniques such as Dictionary Learning (DL)~\cite{mairal2009online} and Non-negative Matrix Factorization (NMF)~\cite{fevotte2011algorithms}. We present a novel coding system named Data-driven Facial Expression Coding System (DFECS) and derive its AUs by employing the Full Face Model (FFM) on the DISFA dataset. Our results indicate that, while both PCA AUs and DFECS AUs exhibit similar performance in terms of variance explained on test datasets (CK+ and BP4D-Spontaneous), the latter demonstrates a notable improvement in interpretability. Specifically, 87.5 percent of DFECS AUs are interpretable, representing a substantial enhancement compared to the 50 percent achieved by the PCA AUs~\cite{chandra2023pca}. In the future, automated keypoint-based learning of facial coding systems with more accurate and stable tracking can pace up facial expression analysis. For example, in security, it might speed up abnormal behavior analysis for forensic investigations; in healthcare, it could aid in managing conditions like pain, depression, or schizophrenia. Similar to existing coding systems~\cite{zhi2020comprehensive,chandra2023pca}, DFECS may be a versatile tool with numerous applications in entertainment, marketing, education, animation, facial expression synthesis, emotion quantification, and robotics.

\par The following section outlines the preprocessing steps before generating facial keypoints and the estimation of DFECS AUs. In Section 3, we delve into the experimentation and present the results, and Section 4 provides a conclusion, including the limitations of our work and potential directions for future research.

\section{Unsupervised learning of DFECS}

\begin{figure}[h]
\centering
\includegraphics[scale=0.27]{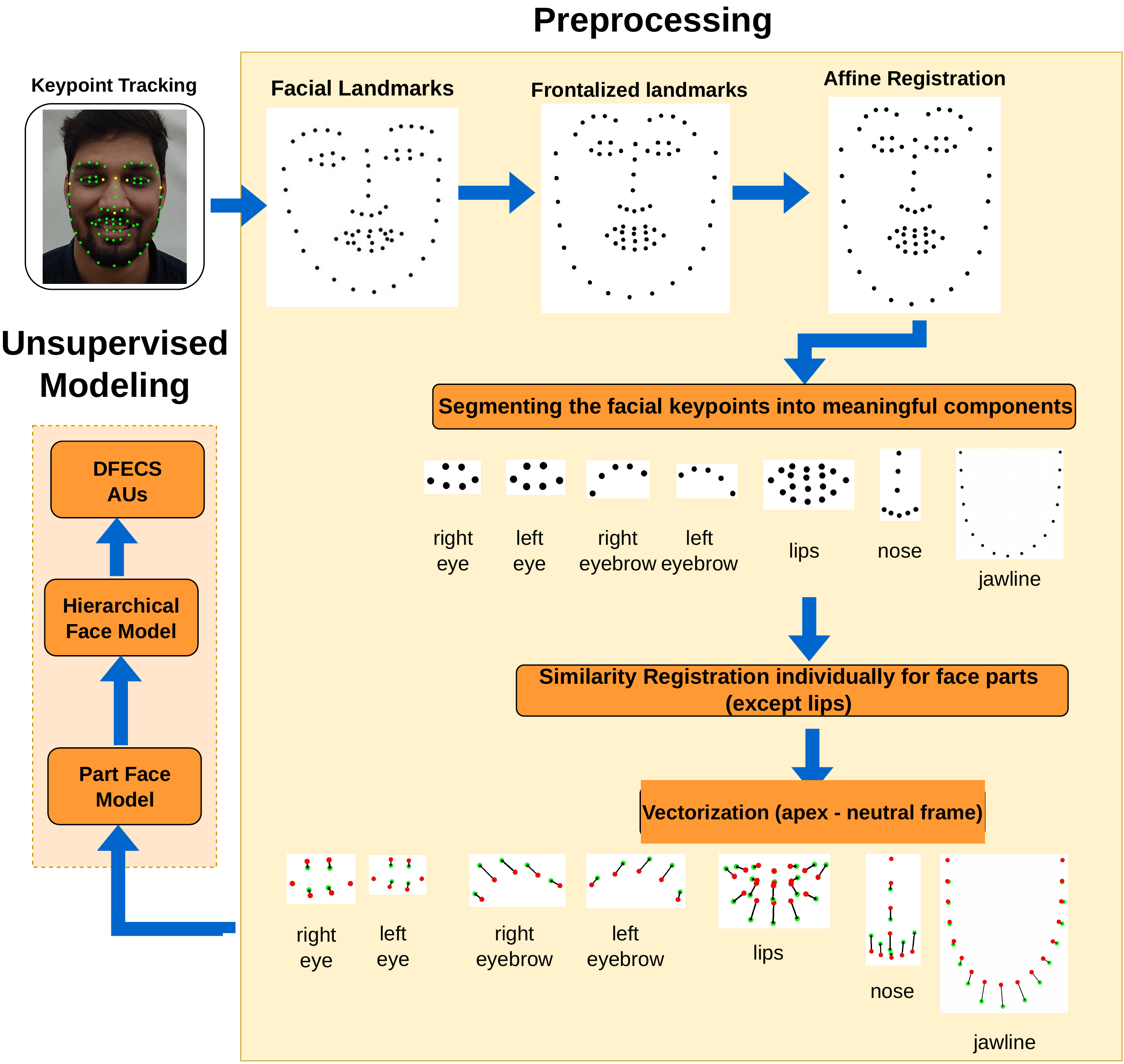}
\caption{Pictorial representation from preprocessing to estimating the final DFECS AUs.}
\label{fig:pipeline}
\end{figure}

\par The essence of facial expressions within video frames can be extracted through tracking the facial keypoints (Fig.~\ref{fig:pipeline}). To eliminate the effect of geometric variabilities across multiple subjects, such as head movement, the difference in face sizes, or relative position of face parts on our model, we first standardize the keypoints as in Tripathi et al.~\cite{chandra2023pca}. The standardized keypoints are finally converted into Keypoint Motion (KPM) vectors representing the changes of facial keypoints from a neutral frame capturing the movements of facial expressions. Finally, we present our FFM model to estimate our DFECS AUs from the KPM dataset. The remainder of this section presents the details of the above steps.

\begin{figure}[h]
    \centering
    \includegraphics[scale=0.25]{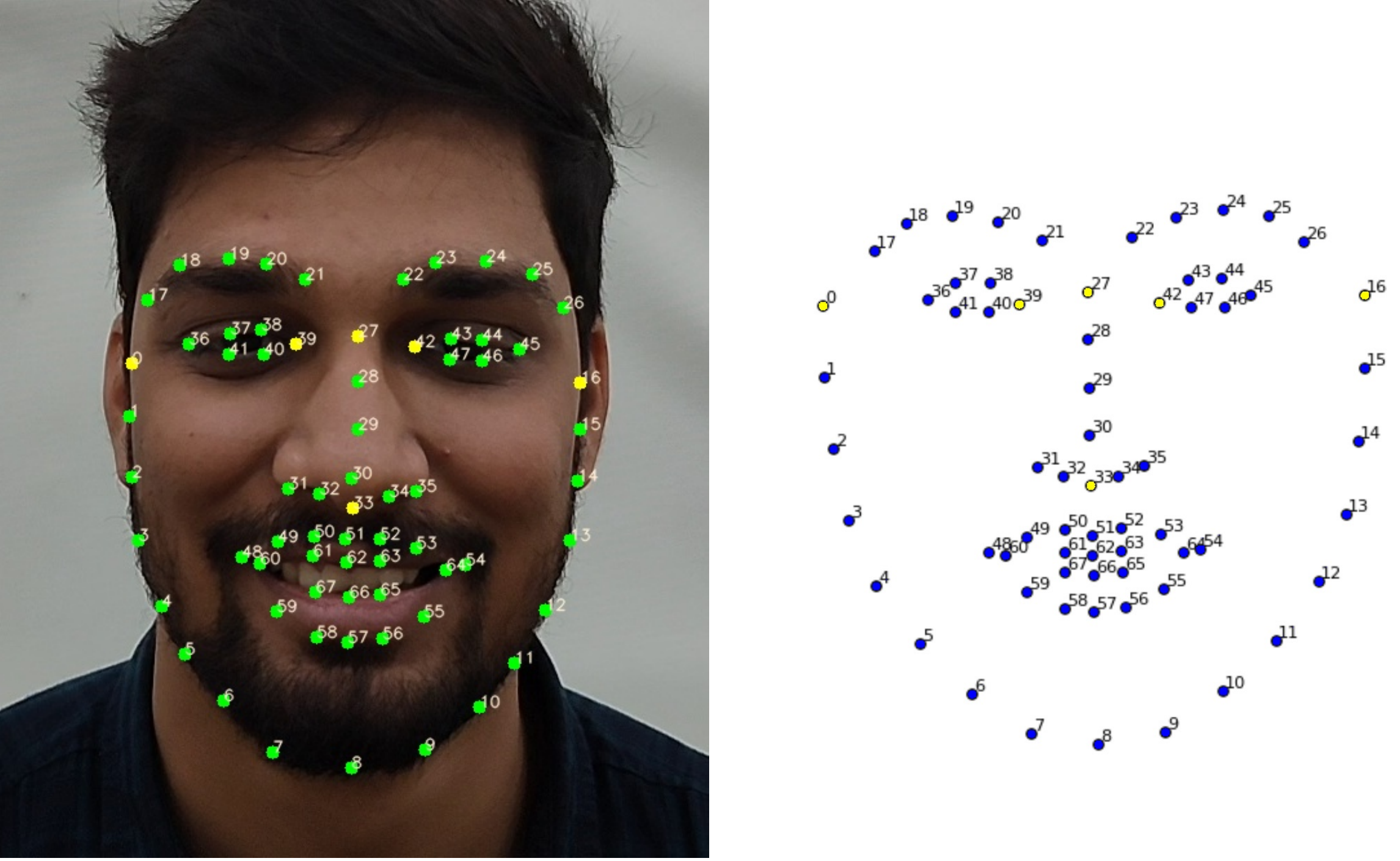}
    \caption{The location of 68 facial keypoints on a subject. Yellow keypoints are employed for registration using affine transformation.}
    \label{fig:datasetskp}
\end{figure}

\subsection{Geometric corrections}
\label{sec:geometric_corrections}
To prevent any potential bias or inaccuracies in our model due to geometric variabilities across different subjects—such as head movement, the difference in face sizes, or the relative position of face parts—we preprocess facial keypoints. The preprocessed keypoints are called standardized keypoints. The process of eliminating geometric variations involves three key steps, outlined below:

\begin{enumerate}[label={}]
    \item \textbf{Facial Keypoints Frontalization: }
	Head movements can obscure pure facial muscle movement in an image. For instance, to quantify a facial muscle movement at a keypoint position, the displacement of the keypoint cannot be used unless there is no head movement. Head movements must be eliminated to obtain accurate facial muscle movements. To achieve this, we employ the algorithm introduced by Vonikakis et al.~\cite{9190989}, which effectively aligns facial keypoints of all the video frames to a front-facing orientation, thereby eliminating keypoint movements due to head motion.

    \item \textbf{Affine Registration: } 
   Subsequently, we perform six-parameter affine registration (translation, rotation, scaling, and shearing)~\cite{hartley2003multiple} to standardize all keypoints into a common space, thereby mitigating variabilities related to face size and position. For this step, we select a specific set of six anchor facial keypoints to estimate the six geometric parameters of affine transformation within a face image. The remaining keypoints in the image are then registered based on these parameters. The chosen anchor keypoints remain nearly stationary on the face regardless of facial expression. Specifically, for the 68-keypoints template, these fixated keypoints are zero (0), sixteen (16), twenty-seven (27), thirty-three (33), thirty-nine (39), and forty-two (42), as shown in yellow color in Fig.~\ref{fig:datasetskp}\footnote{Right image is modified form of \url{https://github.com/Fang-Haoshu/Halpe-FullBody/blob/master/docs/face.jpg}}.

    \item \textbf{Similarity Registration: } 
   Finally, we apply four-parameter similarity registration (translation, rotation, and scaling)~\cite{hartley2003multiple} individually to face parts to further eliminate variabilities, such as the distance between the eye corners, the length of the nose and the eyebrows, among others. The anchor keypoints used to estimate the four geometric parameters are as follows: (42,45) keypoints for the left eyebrow and left eye, (36,39) for the right eyebrow and right eye, (27) for the nose, and (0,16) for the jawline (Fig.~\ref{fig:datasetskp}). While similarity registration typically requires at least two point correspondences~\cite{hartley2003multiple}, the nose, having only one fixated keypoint, is translated to the fixated keypoint without undergoing similarity registration. Moreover, since lips lack fixated keypoints, they are excluded from similarity registration.

\end{enumerate}

\subsection{Keypoint Motion (KPM) vectors}
\label{sec:kpm-generation}

\begin{figure}[h]
    \centering
\includegraphics[scale=0.40]{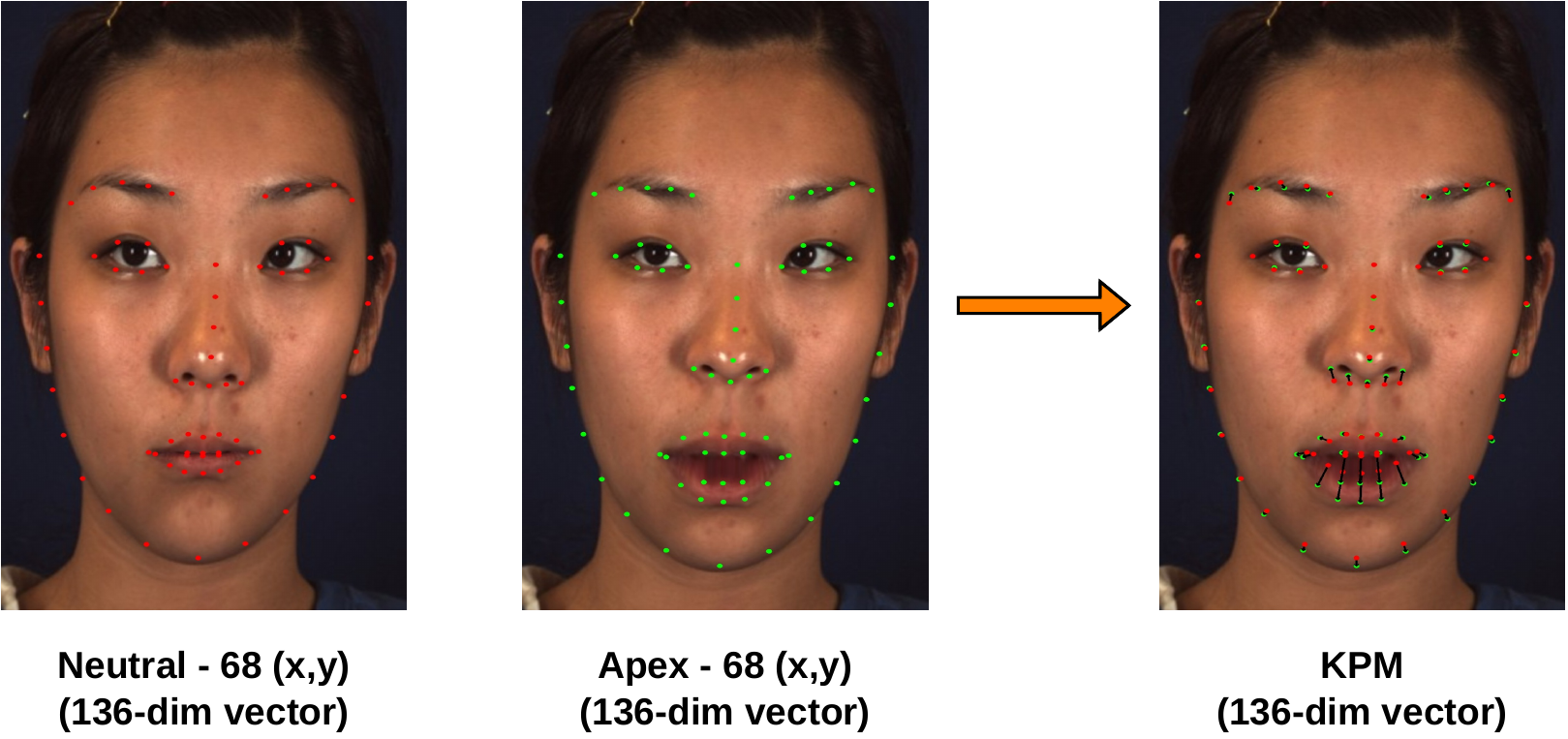}
    \caption{Generation and visualization of a KPM on a face image. Red keypoints indicate the neutral image's facial keypoints, green ones represent the expression frame keypoints, and arrows illustrate how the keypoints move from the neutral position to the corresponding expression image represented by the KPM. (Face image source: BP4D-Spontaneous~\cite{zhang2013high,zhang2014bp4d})}
    \label{fig:sample_KPM}
\end{figure}

We define a Keypoint Motion (KPM) vector as a representation of changes in facial keypoint positions from a neutral face. The dimension of this vector is two times the number of keypoints tracked in the videos, denoting the x and y direction movement for each keypoint. To generate KPM vectors, we calculate the difference between the coordinates (x, y) of the standardized neutral frame keypoints and the corresponding standardized coordinates of those keypoints in all other frames of the same subject (Fig~\ref{fig:sample_KPM}).\footnote{The tool used for facial image morphing ``Pychubby'' is available at: \url{https://github.com/jankrepl/pychubby/}} Consequently, the facial expression at any frame in a video is captured by an r-dimensional vector (where r is twice the dimensions of the number of keypoints tracked), encoding the movements of the keypoints from the subject's neutral frame. To compactly represent the data from all subjects, we utilize a matrix \(X \in \mathbb{R}^{r \times m}\), where $r$ represents the (x, y) coordinate movements of the $\frac{r}{2}$ keypoints from the neutral frame, and \(m\) signifies the sum of the total number of video frames for each subject in the dataset.

\subsection{Estimating DFECS AUs}

To estimate DFECS AUs, we employ an unsupervised algorithm that aims to learn a small set of ``basis vectors'' from a large sample of KPM vectors, which can accurately represent any KPM sample. This task is similar to the classical framework of dimensionality reduction~\cite{sorzano2014survey,van2009dimensionality,reddy2020analysis}. The dimensionality reduction problem may be restated in terms of the matrix factorization problem as follows. Consider m-points (or KPM samples) in a r-dimensional space, represented as a matrix $X\in \mathbb{R}^{r\times m}$. Our objective is to represent these points in a k-dimensional space (generally $k<<r$), using a set of $k$ basis vectors in $\mathbb{R}^{r}$ represented as a matrix $U\in \mathbb{R}^{r\times k}$. These $k$-basis vectors ($u_{j}$ where $j\in \{1,2,...k\}$) represent the data in a much more meaningful manner. Any point $x_{i}$ may thus be approximated as a linear combination of these basis vectors ($x_{i}\approx \sum_{j=1}^{k}u_{j}v_{ij}$, $v_{ij}$ are the linear weights). This may be compactly written in matrix form as $X\approx UV$. The corresponding matrix factorization thus becomes $\min \|X-UV\|$ subject to constraints on $U,V$. $U$ is the component matrix, and its columns are called components. $V$ is the encoding matrix, and its columns are called encodings. The k basis vectors depict the movements of facial keypoints from a neutral facial expression. Like FACS AUs, which can capture any facial expression with only 26 units (excluding head movements), these basis vectors efficiently represent the entire dataset, $X$.

\par The PCA algorithm emerges as a natural choice for generating low-dimensional representations of KPM vectors~\cite{chandra2023pca}. The PCA decomposition problem may be written as $\min \|X-ADB^{T}\|^{2}_{F}$ subject to $A\in \mathbb{R}^{p\times k}, B\in \mathbb{R}^{n\times k}$, where $D$ is a $k\times k$ diagonal matrix and columns of $A$ and $B$ are constrained to be orthonormal. Defining $A$ as $U$ and $DB^{T}$ as $V$, the PCA decomposition can also be represented as $X\approx UV$. However, the PCA-based estimation of AUs suffer from the problem of interpretability due to dense values in the principal components and encoding matrix~\cite{chandra2023pca}.

\par To improve the interpretability of the DFECS system, we propose a Full Face Model (FFM) to estimate DFECS AUs using advanced dimensionality reduction techniques such as dictionary learning (DL) and non-negative matrix factorization (NMF)~\cite{mairal2009online,hoyer2004non,cichocki2009fast,fevotte2011algorithms}. FFM performs a two-level decomposition to the data. We first create partitions of the input data $X\in \mathbb{R}^{r\times m}$ representing the seven face parts- left eyebrow, right eyebrow, left eye, right eye, nose, lips, and jawline. The first level decomposition is applied to these seven parts individually and are called Part Face Models (PFMs) and capture localized movements of the face parts. To capture the correlations of localized face part movements, the decompositions of the seven PFMs are aggregated and decomposed further and is called Hierarchial Face Model (HFM). For simplicity, we call the aggregated PFMs and HFMs decomposition together as Full Face Model (FFM) (Fig.~\ref{fig:pipeline}). Below, we present the details of both the decompositions and how they combine to represent our final FFM model.

\subsubsection{Part Face Model (PFM)}This model applies decomposition individually to the seven face parts ($f$) - left eyebrow, right eyebrow, left eye, right eye, nose, lips, and jawline. Let $p$ denote the number of dimensions in $X$ corresponding to the face part $f$; for example, for the left eyebrow, 5 keypoints are tracked, leading to $p=10\ (5\times 2)$. Let $X_{f}\in \mathbb{R}^{p\times m}$ denote the submatrix of $X$ corresponding to the set of rows in the face part $f$. For each face-part $f$, the submatrix $X_{f}$ is decomposed as $X_{f}\approx U_{f}V_{f}$, $U_{f}\in\mathbb{R}^{p\times k_{f}},V_{f}\in\mathbb{R}^{k_{f}\times{m}}$ (where $k_{f}<<p$) using Dictionary Learning (DL) algorithm~\cite{mairal2009online} which utilizes the positive LASSO objective given by: 

\begin{equation}
\label{eqn:DL-objective}
\underset{U_f, v_i}{\arg\min}\frac{1}{2}\|X_f-U_fV_f\|^{2}_{F}+\alpha\|V_f\|_{1}\ st.\ \|u_j\|_2\le 1,v_{i,j}\ge 0\ \forall i,j
\end{equation}

The basis vectors in $U_{f}$ are expected to represent facial muscle movements, which typically move unidirectionally from a neutral face. Therefore, the DL applies constraints that ensure positive coefficients in $V_{f}$ for the unidirectional movement of basis vectors from a neutral face. Further, the DL intends to enforce sparsity in the part-face encodings $v_{i}$ to capture finer muscle movements; for example, in an eyebrow movement, sometimes, only the inner part is pulled upwards, or only the outer part instead of the full eyebrow.
       
\subsubsection{Hierarchical Face Model (HFM)}
\label{subsec:HFM}
For any face part $f$ ($f$ = 1, 2, ... 7), PFMs are represented as $X_{f}\approx U_{f}V_{f}$ having $k_{f}$ components in its component matrix ($U_{f}$). Each of the p-dimensional PFM component in $U_{f}$ is expanded to a full face ($\frac{r}{2}$ facial keypoints) template, i.e., into r-dimensional vector (by adding a zero to keypoints not present in the corresponding face part). The $U_{f}$ corresponding to the seven face parts are then stacked together in the matrix $U\in \mathbb{R}^{r\times k}$, where $k=\sum_{f=1}^{7}k_{f}$. Similarly, $V_{f}$ are stacked vertically in a matrix $V\in \mathbb{R}^{k\times m}$. Then, the PFM decomposition can be compactly represented as $X\approx UV$.
\par In reality, the movement of keypoints across different face parts are expected to be correlated (e.g., movement of the left eye is expected to accompany movement of the left eyebrow). To capture correlations of the keypoint movements in different face parts, we apply a 2nd level decomposition called the Hierarchical Face Model (HFM). HFM decomposition is applied on the matrix $V\ (v_{i,j}\ge 0\ \forall i,j)$, such that $V\approx AB$, $A\in\mathbb{R}^{k\times q},B\in\mathbb{R}^{q\times m}$ (where $q<<k$) using modified NMF algorithm~\cite{hoyer2004non,cichocki2009fast,fevotte2011algorithms} which solves the objective:

\begin{equation}
\label{eqn:NMF-objective}
\underset{A, B}{\arg\min}\|V-AB\|^{2}_{F}+\alpha_A\|A\|_{1}+\alpha_B\|B\|_{1}\ st.\ A\ge0\, B\ge0
\end{equation}

The positivity of $A$ ensures positively correlated movements of facial parts; for instance, in a happy expression, lip corner pulling often accompanies cheek raising and not lowering. The positivity of $B$ ensures the unidirectional movement of facial keypoints from a neutral position in the correlated part face components. L1 regularization on $A$ enforces sparsity intended to capture correlated face part components in which only few face parts are moving, for example, a mild smile may only involve the movement of brows, eyes and lips. Similarly, L1 regularization on $B$ is enforced to allow the possibility of generating sparse codings where only a few parts are moving. 

\par The final model termed FFM can be written as $X\approx \hat{X}=UAB$. $U$ represents the part face components and $A$ captures the correlations between them. Therefore, our final DFECS AUs estimated by the FFM model are compactly represented by $U^{'}=UA$, and $B$ represents the encoding matrix.

\subsubsection{Hyperparameter Tuning for learning FFM}

In this section, we provide the details of hyperparameter tuning for both PFM and HFM to train an FFM model. The model learning aims to achieve a specific given $(1-\beta)$ percent of variance explained in the final FFM model. Note that FFM performs dimensionality reduction using matrix decomposition twice, first for PFM and then for HFM. Each decomposition introduces some error due to reduced dimensions. Consequently, the PFM decomposition aims to attain an explained variance higher than $(1-\frac{\beta}{2})$ so that the subsequent error, resulting from reduced dimensions during the HFM, can reach a variance explained percentage of $(1-\beta)$.

\par For the PFM represented as $\hat{X}_f \approx U_fV_f$, two hyperparameters are tuned: the number of face part components $k_f$ and the sparsity controller ($\alpha$)(Eq.~\ref{eqn:DL-objective}) of the encoding matrix $V_f$. Training is halted when the percent variance explained reaches $(1-\frac{\beta}{2})$. Grid search is conducted for each PFM, where rows signify the variation of $k_f$ from $k_f=1$ to $k_f=p$ (where $p$ denotes the number of dimensions in the face part $f$), and columns signify the variation of the sparsity constraint from $\alpha=5$ to $\alpha=0.5$ with a decrement step of $0.5$.

\par For the HFM, which utilizes the NMF algorithm, three hyperparameters are tuned: the number of components $q$, the sparsity controller of components $\alpha_{A}$, and the sparsity controller of the encoding matrix $\alpha_{B}$ (Eq.~\ref{eqn:NMF-objective}). NMF training employs a three-dimensional grid search, where the first dimension covers $q=1$ to $q=k$ ($k=\sum^{f=7}_{f=1} k_f$, is the total number of face part components across all face parts), the second dimension varies $\alpha_{A}$ from $\alpha_{A}=5$ to $\alpha_{A}=0.5$ with a decrement step of 0.5, and the third dimension varies $\alpha_{B}$ from $\alpha_{B}=5$ to $\alpha_{B}=0.5$ with a decrement step of 0.5. The training of HFM ($V\approx AB$) concludes when the percent variance explained (between $X$ and $UAB$, refer section~\ref{subsec:HFM} and ~\ref{subsec:metrics}) reaches $(1-\beta)$.

\section{Evaluation methodology}
\label{sec:evaluation_methodology}

The FFM learns a decomposition of the input KPMs (represented as matrix $X$) as $X\approx UAB$, where $U\in\mathbb{R}^{r\times k}, A\in\mathbb{R}^{k\times q},B\in\mathbb{R}^{q\times m}$. The columns of matrix $U^{'}=UA$ represent the different DFECS AUs learned by the FFM. To compare this AU representation with FACS, one needs to (a) generate corresponding component matrix $U_{FACS}$ for FACS AUs, (b) The encoding matrix $V_{FACS}$, such that $X\approx U_{FACS}V_{FACS}$. The next subsection describes the methods used to generate the component matrix for FACS AUs and the encoding matrix to compare DFECS AUs, FACS AUs, and PCA AUs. We further outline the datasets and the evaluation metrics used.

\subsection{Computing component matrix for FACS AUs}
\label{sec:FACS_as_KPMs}

We first categorize FACS AUs into two subsets: pure AUs, representing pure atomic expressions, and comb AUs, encompassing both pure AUs and some of their combinations mentioned in the FACS manual~\cite{ekman1978facial, ekman2002facial}. To isolate pure facial muscle movements, we identify 26 pure AUs and 113 comb AUs from the AUs specified in the FACS manual, excluding any AUs related to head movement. Next, we focus on a single subject and extract images for the expression (apex) frame of each pure AU and comb AU from the FACS manual~\cite{ekman2002facial}. Additionally, a neutral image of the same subject is selected. Using a keypoint tracking algorithm~\cite{dong2020supervision}, we generate 68 facial keypoints on each image. These keypoints then undergo the same preprocessing steps outlined in sections~\ref{sec:geometric_corrections} and ~\ref{sec:kpm-generation}, resulting in a standardized representation of the 136-dimensional KPM vector for each AU. The compiled KPM vectors for the 26 pure AUs are denoted in a matrix \(U_{FACS,1} \in \mathbb{R}^{136 \times 26}\), while those for the 113 comb AUs in a matrix \(U_{FACS,2} \in \mathbb{R}^{136 \times 113}\).

\subsection{Encoding matrix computation}
\label{subsec:encoding_matrix_computation}

Given a KPM dataset, represented as a matrix $X\in\mathbb{R}^{r\times m}$ and an action unit (DFECS AUs, PCA AUs or FACS AUs) represented using a matrix $U\in\mathbb{R}^{r\times k}$, our goal is to compute the encoding matrix $V^{''}\in\mathbb{R}^{k\times m}$ such that $UV^{''}$ approximates $X$ ($X\approx UV^{''}$). For the DFECS AUs the encoding matrix is constrained to be non-negative. This is obtained by solving the following optimization problem: 

\begin{equation}
\label{Unsupervised:encoding_objective}
\underset{V^{''}}{\arg\min}\|X-UV^{''}\|^{2}_{F}+\alpha\|V^{''}\|_{1}
\end{equation}
s.t. $v^{''}_{ij}\ge 0$ for all $i\in 1,2,....,k$ and $j\in 1,2,....,m$. The parameter $\alpha$ and the L1-norm penalty are used in the objective to promote and control sparsity in the encoding matrix $V^{''}$. The sparsity ensures the presence of fewer AUs in the coding of a sample where only a few facial muscles are moving. The above problem can be decomposed into $m$ independent identical problems on different KPM samples as follows:

\begin{equation}
  \min \|y_{i}-Uv^{''}_{i}\|^{2}_{2}+\alpha\|v^{''}_{i}\|_{1}, s.t. \forall v^{''}_{ij}\ge 0
\end{equation}

The above problem may be solved using positive lasso~\cite{efron2004least}, which also gives the complete solution path with different solutions for different values of $\alpha$.

\par Note that there was no constraint on the encoding matrix for PCA AUs in Tripathi et al.~\cite{chandra2023pca}, which contained both positive and negative values. Therefore, PCA AUs were allowed to move in positive as well as negative directions. For a fair comparison with DFECS AUs, we must also learn positive encodings for PCA AUs. However, upon constraining the encoding matrix of PCA AUs to be positive values only, we need to include the negative of the components in the PCA AUs to represent the original set. Therefore, the component matrix $U$ of PCA AUs is updated as $U=[U\ -U]$.

\subsection{Evaluation datasets}

To evaluate the performance of our FFM model, we use three datasets: CK+~\cite{kanade2000comprehensive,lucey2010extended}, DISFA~\cite{mavadati2012automatic,mavadati2013disfa}, and BP4D-Spontaneous~\cite{zhang2013high,zhang2014bp4d} described next.\\

\textbf{CK+:} This dataset comprises 593 posed facial video recordings of 123 subjects (aged 18-50) recorded at 30 frames per second (FPS). It captures facial expressions displaying emotions such as anger, contempt, disgust, fear, happiness, sadness, and surprise. The dataset contains around 10,100 video frames, resulting in a data matrix $X$ of size $10,100\times 136$.\\

\textbf{DISFA:} In this dataset, there are spontaneous facial expression videos of 27 subjects (aged 18-50) recorded at 20 FPS. These expressions are elicited using a 4:01 minute emotionally evocative video targeting five distinct emotions: disgust, fear, happiness, sadness, and surprise. The dataset comprises 1,14,000 video frames, generating the matrix $X$ of size $1,14,000\times 136$. To ensure a fair comparison with CK+, we randomly sample 10,100 KPMs from DISFA for training, denoting it as DISFA (train).\\

\textbf{BP4D-Spontaneous:} This dataset features spontaneous facial expression videos of 41 subjects (aged 18-29) recorded at 25 FPS. Expressions are elicited through eight different activities, including interview, watching a video clip, startle probe, improvisation, threat, cold pressor, insult and smelling. A total of 1,36,000 frames in this dataset results in a data matrix $X$ of size $1,36,000\times 136$. For a fair comparison with training on CK+ and DISFA, we randomly sample 10,100 KPMs from BP4D and designate it as BP4D (train). Note that, in the case of BP4D-Spontaneous, all the keypoints on the jawline are missing (Fig.~\ref{fig:datasetskp}). Therefore, the affine registration step when converting the tracked facial keypoints to standardized keypoints utilizes the following anchor points: twenty-seven (27), thirty-three (33), thirty-nine (39), forty-two (42), thirty-six (36), and forty-five (45) were used.\\

\par These datasets come pre-labeled with facial keypoints. These keypoints were tracked using different computer vision techniques: Active Appearance Model (AAM)~\cite{cootes2001active} for DISFA (66 keypoints), Active Appearance Model (AAM)~\cite{cootes2001active,matthews2004active} for CK+ (68 keypoints), and the Constrained Local Model (CLM)~\cite{saragih2009deformable,saragih2011deformable} for BP4D-Spontaneous (49 keypoints). To ensure uniform representation, all datasets are vectorized into 68 keypoint templates, with missing keypoints being filled with zero values. Further, incorporating spontaneous datasets like DISFA and BP4D-Spontaneous enables us to capture a wide range of facial expressions, enhancing the model's ability to generalize. Additionally, we include the dataset CK+ containing posed expressions and spontaneous smiles to investigate potential improvements in the final FFM model.\\

\subsection{Evaluation metrics}
\label{subsec:metrics}

To assess the performance of our FFM model, we adopt and modify some of the metrics proposed by Tripathi et al.~\cite{chandra2023pca}, which were originally designed for evaluating the PCA model performance. These have been presented below.

\begin{enumerate}

\item \textbf{Train Variance Explained (VE (train)) - }Consider a model trained on the dataset \(X\), where \(X \in \mathbb{R}^{136 \times m}\). VE (train) measures how well the DFECS AUs reconstruct the training data and is defined as the percentage of variance explained, calculated as \(100 \times \left(1 - \frac{\| X - \hat{X} \|_{F}^{2}}{\| X \|_{F}^{2}}\right)\), where \(\hat{X}\) is the low-dimensional representation of \(X\) given by \(UV^{'}\).

\item \textbf{Test Variance Explained (VE (test)) - } Given a test dataset \(Y\) and AU component matrix $U$, we compute its encoding matrix (section~\ref{subsec:encoding_matrix_computation}) such that $Y\approx \hat{Y}=UV^{''}$. VE (test) assesses how accurately the AU matrix ($U$) reconstructs the test data. It is defined as the percentage of variance explained, computed as \(100 \times \left(1 - \frac{\| Y - \hat{Y} \|_{F}^{2}}{\| Y \|_{F}^{2}}\right)\).

\item \textbf{Interpretability - } The \textit{interpretability} metric evaluates the degree of biologically plausible movements within the DFECS AUs. Firstly, all DFECS AUs are visually depicted by projecting them onto a neutral face~\cite{chandra2023pca}. This involves selecting facial keypoints from a neutral image and then adding a 136-dimensional KPM vector representing a DFECS AU to these keypoints, resulting in an expressive frame. Facial components are then morphed to align with the expression's keypoints, yielding a projected AU image (see SI Fig. 1). Three independent volunteers then examine each keypoint movement in any DFECS AU projected image by comparing it with the feasible directions of muscle movements depicted in Fig.~\ref{musclefig1} at that keypoint location. If any keypoint movement within any AU deviates from the possible movements of facial muscles in that facial area, the volunteer labels it as a non-interpretable AU. Finally, we apply majority voting, where a DFECS AU is considered non-interpretable if at least two of the volunteers label it as non-interpretable. Let $ni$ denote the number of non-interpretable AUs from the total $k$ AUs in a DFECS model. Then, the \textit{interpretability} metric is defined as $\frac{k-ni}{k}\times 100$.

\begin{figure}[h]
    \centering
\includegraphics[scale=0.20]{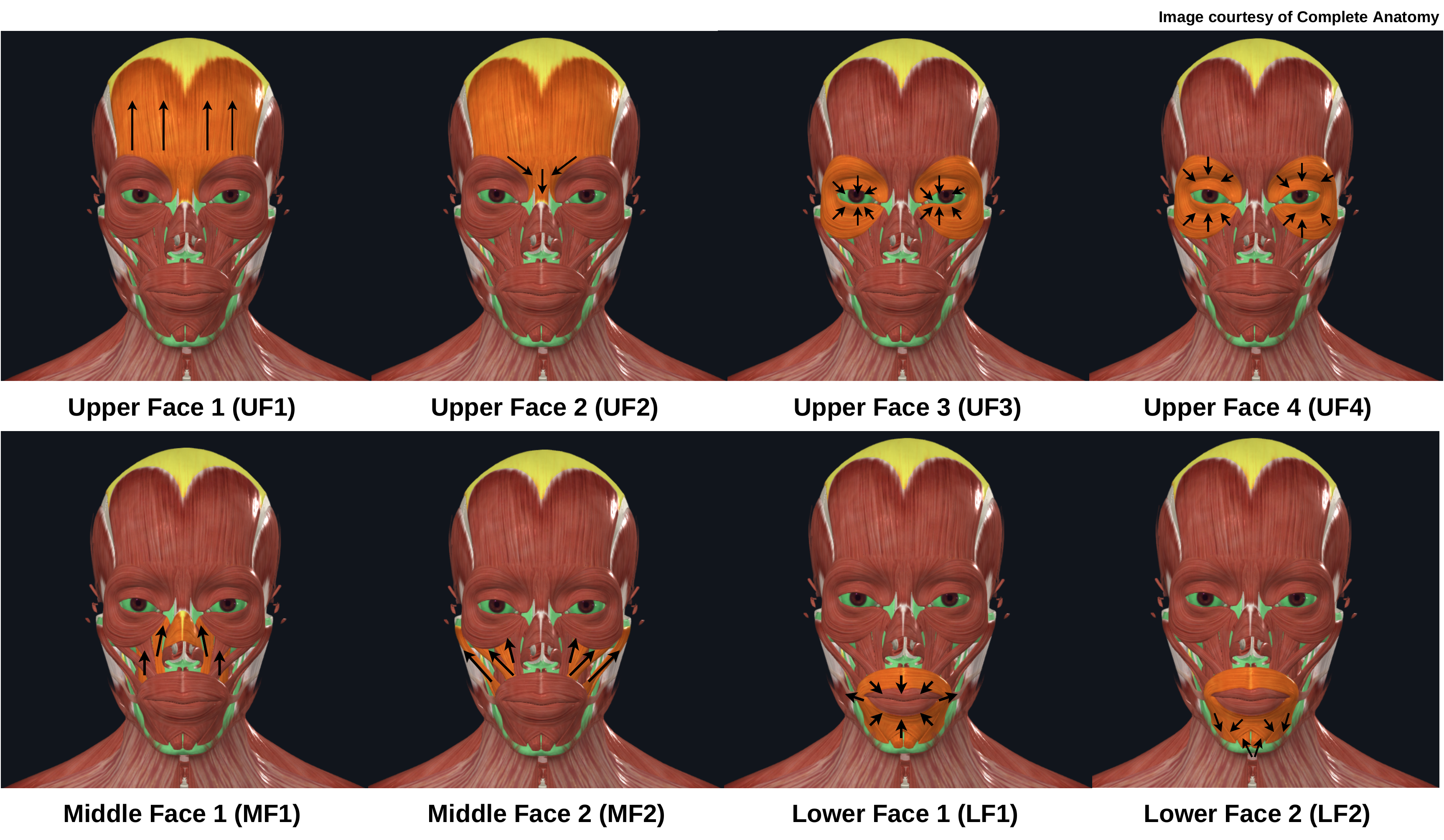}
    \caption{Muscle movements of AUs in Facial Action Coding System (FACS) excluding AUs coding for head movements.}
    \label{musclefig1}
\end{figure}

\end{enumerate}

\section{Results}
In this section, we first compare the DFECS AUs generated by the three training datasets (DISFA (train), BP4D (train), and CK+) for their coding power and interpretability using the evaluation metrics. We finally compare the DFECS AUs with PCA AUs and the component matrix of FACS AUs.

\subsubsection{Variance Explained}

\begin{figure}[h]
    \centering
    \includegraphics[scale=0.16]{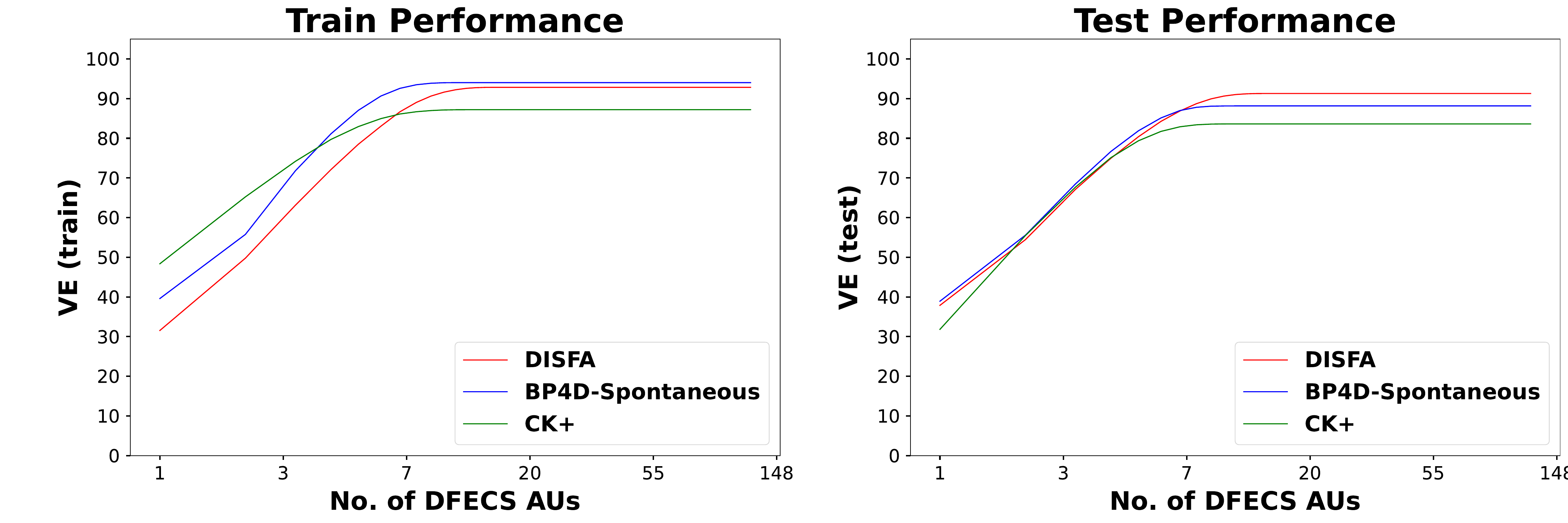}
    \caption{VE (train) and the average of the VE (test) on remaining datasets when DFECS AUs trained on DISFA, CK+, and BP4D-Spontaneous are utilized to generate the positive encoding matrices on train and test datasets. (x-axis on the log scale)}
    \label{fig:L0_train_performance}
\end{figure}

We first compare the train and test performance of DFECS AUs on different training and testing datasets. For this, an FFM model was first learned on a train dataset, then the encoding matrices for train and different test datasets were computed (section~\ref{subsec:encoding_matrix_computation}). The hyperparameters for the final FFM model that achieved a Var (train) of 95 percent when trained on the DISFA (train), BP4D (train), and CK+ datasets can be found in the SI Table. 1 and SI Table. 2.

For a given training dataset, and the value $k$, representing the number of AUs utilized by a sample, we compute the mean of VE (train) over all samples and the mean VE (test) across all samples for the test datasets. Subsequently, for a given train dataset, at each $k$, we take the average VE (test) over the remaining two test datasets.
The results are shown in Fig.~\ref{fig:L0_train_performance} for the datasets- DISFA, CK+, and BP4D-Spontaneous. It's important to note that, on the x-axis of the graph, each corresponding VE (train) and VE (test) for a given $k$ signifies the variance explained with $k$ or fewer than $k$ components. This is due to the fact that not every sample can utilize up to $k$ AUs, owing to the constraints imposed by positive encodings within the subspace spanned by the AUs. 

\par The average VE (test) across the three datasets remains comparable when a small number of AUs is utilized ($k<7$). However, asymptotically, DFECS AUs estimated from DISFA exhibit the best average VE (test). 
Consequently, we chose the FFM components from DISFA as the final DFECS AUs, comprising sixteen components in the component matrix $U$ and having Var (train) of 95 percent.

\subsubsection{Interpretability}

\par
\begin{figure}[htp]
    \centering
    \includegraphics[scale=0.06]{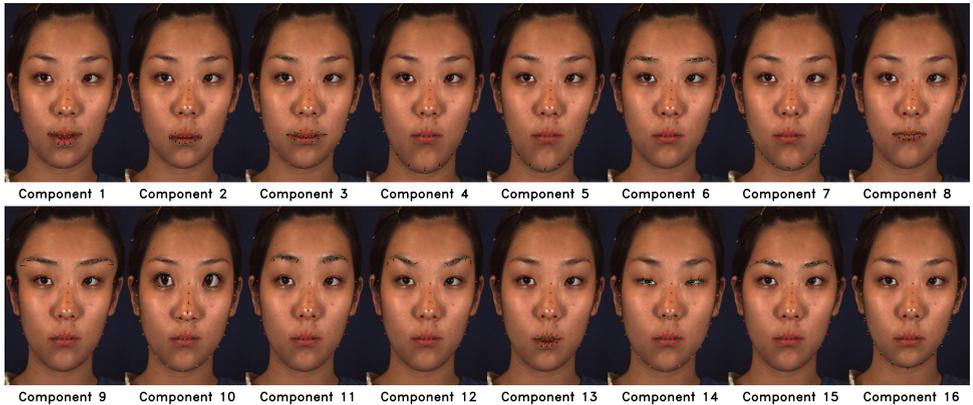}
    \caption{Visualization of sixteen DFECS AUs estimated from DISFA (train). (Face image source: BP4D-Spontaneous~\cite{zhang2013high,zhang2014bp4d})}
    \label{fig:FFM_AUs}
\end{figure}

 We further inspect the \textit{interpretability} metric for the sixteen components of DFECS AUs. These components represented as KPM vectors, are visualized by projecting them onto a neutral face (see Fig.~\ref{fig:FFM_AUs}). The labeling of these AUs into interpretable or non-interpretable categories by the three volunteers (see section~\ref{subsec:metrics}) is shown in Table~\ref{tab:FFM_interpretability}. The majority vote reveals that Component 6 and Component 9 are non-interpretable; therefore, 14 out of the 16 components are fully interpretable, or the \textit{interpretability} metric is 87.5 percent. For demonstration purposes, the non-interpretable keypoint movements coded by volunteer 1 are depicted as blue dots in Fig.~\ref{fig:FFM_AUs} and detailed in SI Table. 4.

\begin{table}[!htp]
\centering
\caption{Interpretability labeling of DFECS AUs by the three volunteers. A volunteer marks any AU as either interpretable (\checkmark) or non-interpretable (\texttimes).}
\label{tab:FFM_interpretability}
\scriptsize
\begin{tabular}{lcccc}
\toprule
\textbf{Component} & \textbf{Volunteer 1} & \textbf{Volunteer 2} & \textbf{Volunteer 3} & \textbf{Majority Vote} \\
\midrule
Component 1 & \checkmark & \checkmark & \checkmark & \checkmark \\
Component 2 & \checkmark & \checkmark & \checkmark & \checkmark \\
Component 3 & \checkmark & \checkmark & \checkmark & \checkmark \\
Component 4 & \checkmark & \checkmark & \checkmark & \checkmark \\
Component 5 & \checkmark & \texttimes & \checkmark & \checkmark \\
Component 6 & \texttimes & \checkmark & \texttimes & \texttimes \\
Component 7 & \checkmark & \checkmark & \checkmark & \checkmark \\
Component 8 & \checkmark & \checkmark & \checkmark & \checkmark \\
Component 9 & \texttimes & \texttimes & \texttimes & \texttimes \\
Component 10 & \checkmark & \checkmark & \checkmark & \checkmark \\
Component 11 & \checkmark & \checkmark & \checkmark & \checkmark \\
Component 12 & \checkmark & \checkmark & \texttimes & \checkmark \\
Component 13 & \checkmark & \checkmark & \checkmark & \checkmark \\
Component 14 & \checkmark & \checkmark & \checkmark & \checkmark \\
Component 15 & \checkmark & \checkmark & \checkmark & \checkmark \\
Component 16 & \checkmark & \checkmark & \checkmark & \checkmark \\
\midrule
\textbf{Percentage} & \textbf{87.5\%} & \textbf{87.5\%} & \textbf{81.25\%} & \textbf{87.5\%} \\
\bottomrule
\end{tabular}
\end{table}

\subsection{Comparison of PCA AUs, DFECS AUs, and FACS AUs}

\subsubsection{Comparing VE (test)}

\begin{figure}[h]
    \centering
    \includegraphics[scale=0.16]{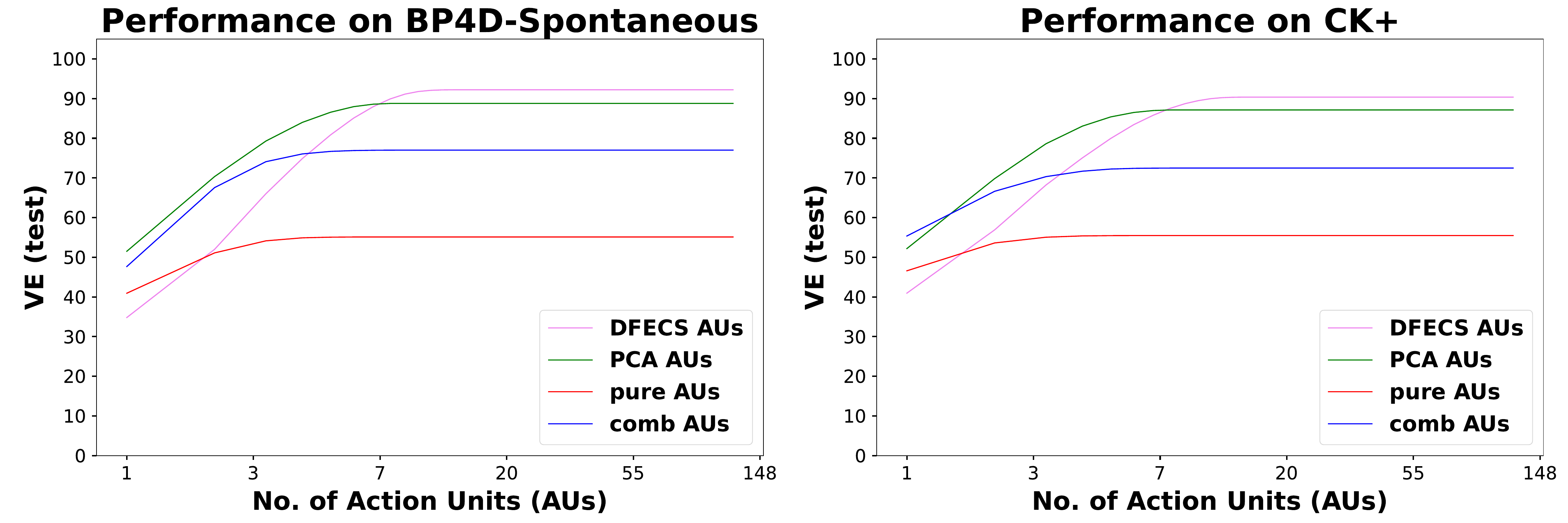}
    \caption{Performance of PCA AUs, DFECS AUs, pure AUs, and comb AUs on the test datasets - CK+ and BP4D-Spontaneous. (x-axis on the log scale)}
    \label{fig:all_aus_L0_compare}
\end{figure}

For a given test data $Y$ and AUs in the component matrix $U$, we compute the encoding matrix (see section~\ref{subsec:encoding_matrix_computation}) such that $Y=\hat{Y}\approx UV^{''}$. Further, at each $k$ representing the number of AUs, where $k$ ranges from 1 to 136, we calculate the mean of the VE (test) across all samples. The results for both CK+ and BP4D-Spontaneous test datasets are shown in Fig.~\ref{fig:all_aus_L0_compare}.

\par In both the BP4D-Spontaneous and CK+ datasets, PCA AUs outperform DFECS AUs and FACS AUs (pure AUs and comb AUs) when a small number of AUs are utilized ($k<7$). Asymptotically, DFECS AUs exhibit the best performance. Additionally, we assess the performance with respect to the L1-norm of the encoding matrix for $L1=0$ to $L1=1000$, computing the mean of the VE (test) across all samples at each L1 value (SI Fig. 2). The results reveal that PCA AUs and DFECS AUs demonstrate comparable performance, and consistently outperform pure AUs and comb AUs.

\subsubsection{Comparing Interpretability}

DFECS AUs demonstrate a significant improvement in interpretability compared to the top eight PCA AUs, which only achieved 50 percent \textit{interpretability} in Tripathi et al.~\cite{chandra2023pca}. However, since the encoding matrix of PCA includes both positive and negative values, while our FFM's encoding matrix comprises only positive values, we analyze the new component matrix for PCA AUs (see section~\ref{subsec:encoding_matrix_computation}) for interpretability for a fair comparison. This results in PCA AUs with sixteen components.

\begin{table}[!htp]
\centering
\caption{Interpretability labeling of PCA AUs by the three volunteers. A volunteer marks any AU as either interpretable (\checkmark) or non-interpretable (\texttimes).}
\label{tab:PCA_interpretability}
\scriptsize
\begin{tabular}{lcccc}
\toprule
\textbf{Component} & \textbf{Volunteer 1} & \textbf{Volunteer 2} & \textbf{Volunteer 3} & \textbf{Majority Vote} \\
\midrule
Component 1 (+ve) & \checkmark & \checkmark & \checkmark & \checkmark \\
Component 2 (+ve) & \checkmark & \checkmark & \checkmark & \checkmark \\
Component 3 (+ve) & \texttimes & \checkmark & \checkmark & \checkmark \\
Component 4 (+ve) & \checkmark & \checkmark & \checkmark & \checkmark \\
Component 5 (+ve) & \checkmark & \checkmark & \checkmark & \checkmark \\
Component 6 (+ve) & \texttimes & \checkmark & \texttimes & \texttimes \\
Component 7 (+ve) & \texttimes & \texttimes & \texttimes & \texttimes \\
Component 8 (+ve) & \texttimes & \texttimes & \texttimes & \texttimes \\
Component 1 (-ve) & \checkmark & \checkmark & \checkmark & \checkmark \\
Component 2 (-ve) & \texttimes & \texttimes & \texttimes & \texttimes \\
Component 3 (-ve) & \checkmark & \checkmark & \checkmark & \checkmark \\
Component 4 (-ve) & \checkmark & \checkmark & \checkmark & \checkmark \\
Component 5 (-ve) & \texttimes & \checkmark & \texttimes & \texttimes \\
Component 6 (-ve) & \texttimes & \checkmark & \texttimes & \texttimes \\
Component 7 (-ve) & \checkmark & \texttimes & \checkmark & \checkmark \\
Component 8 (-ve) & \checkmark & \checkmark & \texttimes & \checkmark \\
\midrule
\textbf{Percentage} & \textbf{56.25\%} & \textbf{75.00\%} & \textbf{56.25\%} & \textbf{62.5\%} \\
\bottomrule
\end{tabular}
\end{table}

\begin{figure}[htp]
    \centering
    \includegraphics[scale=0.06]{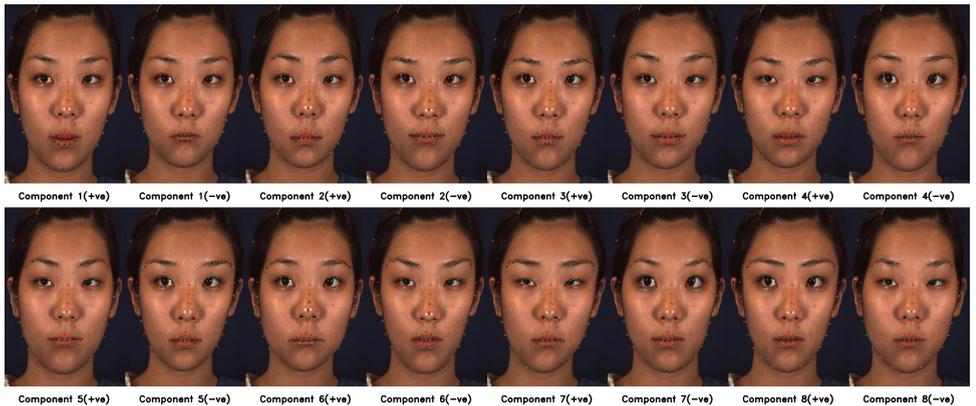}
    \caption{Visualization of the PCA AUs estimated from DISFA (train). The visualization of the components and their corresponding negated components are shown ($U^{new}_{PCA\ AUs}$). Blue dots represent non-interpretable movements. (Face image source: BP4D-Spontaneous~\cite{zhang2013high,zhang2014bp4d})}
    \label{fig:pcaAUs_both}
\end{figure}

\par We examined the \textit{interpretability} metric of sixteen components of PCA AUs, that account for 95 percent VE (train). The interpretability labeling by the three volunteers and their majority vote is presented in Table~\ref{tab:PCA_interpretability}. Component 2 (-ve), Component 5 (-ve), Component 6 (+ve), Component 6 (-ve), Component 7 (+ve), and Component 8 (+ve) of PCA AUs are non-interpretable. Consequently, 10 out of 16 PCA AUs are fully interpretable, or the \textit{interpretability} metric is 62.5 percent. For demonstration purposes, the non-interpretable keypoints movements identified by volunteer 1 are indicated by blue dots in Fig.~\ref{fig:pcaAUs_both} and detailed in SI Table. 3. 

\par Recall that KPM representation of FACS AUs, being directly derived from human facial expressions, have 100 percent \textit{interpretability}. Overall, we find that DFECS AUs exhibit superior performance in explaining variance compared to FACS AUs. However, they are less interpretable compared to KPM-based FACS AUs. Notably, the \textit{interpretability} of DFECS AUs stands at 87.5 percent, a significant improvement over the 62.5 percent \textit{interpretability} of PCA AUs, while having variance explained comparable to that of the PCA AUs.

\section{Conclusion}

In this work, we propose an automated approach to discovering a novel facial coding system DFECS using advanced algorithms (NMF, DL), and estimate its AUs from facial expression videos labeled with keypoints. The DFECS AUs estimated with the DISFA dataset demonstrate comparable variance to PCA AUs and superior variance compared to KPM-based FACS AUs in the BP4D-Spontaneous and CK+ test datasets. Notably, 87.5 percent of DFECS AUs are interpretable, a substantial improvement over the 62.5 percent interpretability achieved with a PCA model in the previous work~\cite{chandra2023pca}. Furthermore, the utilization of positive encodings enhances the interpretation of AUs as facial muscle movements, which typically move unidirectionally from a neutral face.

\par The applications of an automated data-driven facial coding system such as DFECS can be extensive and impactful across various domains. For instance, in the realm of security, such a system can pace up abnormal behavior analysis of facial and gesture behaviors, aiding in forensic investigations. In clinical settings, the system's capabilities in automatic pain analysis and long-term monitoring can offer valuable tools for medical professionals, such as in mental state research related to depression, schizophrenia, and other psychosomatic disorders. Entertainment experiences can be greatly enriched through affect-based interaction, tailoring evoked expressions to users' emotions such as boredom, depression, and happiness. Overall, the automated facial coding system presents a new tool with diverse applications in security, healthcare, entertainment, commerce, education, and robotics, bringing innovation and efficiency to various real-world scenarios.

\par However, certain limitations of the DFECS AUs should be acknowledged. Firstly, we have utilized only 68 facial keypoints, which may limit the system's ability to capture subtle expressions such as wrinkles and microexpressions. Future improvements could involve incorporating more keypoints~\cite{wang2021deep,liu2019grand}, thereby enabling the system to capture a broader range of subtle expressions. Moreover, future efforts should focus on refining the accuracy of keypoint-tracking algorithms, as precise tracking is crucial for capturing subtle expressions such as microexpressions.

Secondly, our interpretability analysis maybe limited because facial expressions are complex and can allow for large number of possible movements, therefore, robust metrics for gauging interpretability can be developed. 
Finally, comparing DFECS AUs with KPM-based FACS AUs may not be the best method. Original FACS offers a qualitative, thorough description of facial muscle movements, allowing to code AUs universally in any expression. Hence, KPM-based FACS AUs were expected to have high coding power. However, they demonstrate lower coding power than DFECS AUs in terms of variance explained in datasets like CK+ and BP4D-Spontaneous. Creating KPM-based FACS AUs using responses from multiple subjects and comparing them with DFECS AUs may be a topic for future exploration.

\section{Funding}
This work was supported by the project “Machine Learning Model for Early Diagnosis of Stroke in Resource Limited Settings” funded by the Department of Biotechnology, Government of India. Grant No.: BT/PR33179/AI/133/16/2019

\newpage
\newgeometry{
left=10mm,
right=10mm,
}


\bibliography{sn-article}





\setcounter{figure}{0} 
\setcounter{table}{0} 

\section{Supplementary material}

\subsection{Generation of a projected AU image}
\label{subsec:projected_AU}

A neutral image, along with its 136-dimensional facial keypoints (comprising 68 x and y coordinates), is chosen. Next, a DFECS AU represented as a KPM vector is added to the 136-dimensional keypoints of the neutral image, resulting in projected AU keypoints. These keypoints are then utilized to morph the neutral image into the projected AU image. The below figure shows a projected AU image. The KPM image in the figure below is an equivalence of the projected AU image with neutral keypoints and direction of movement marked on it.

\begin{figure}[h]
    \centering
\includegraphics[scale=0.40]{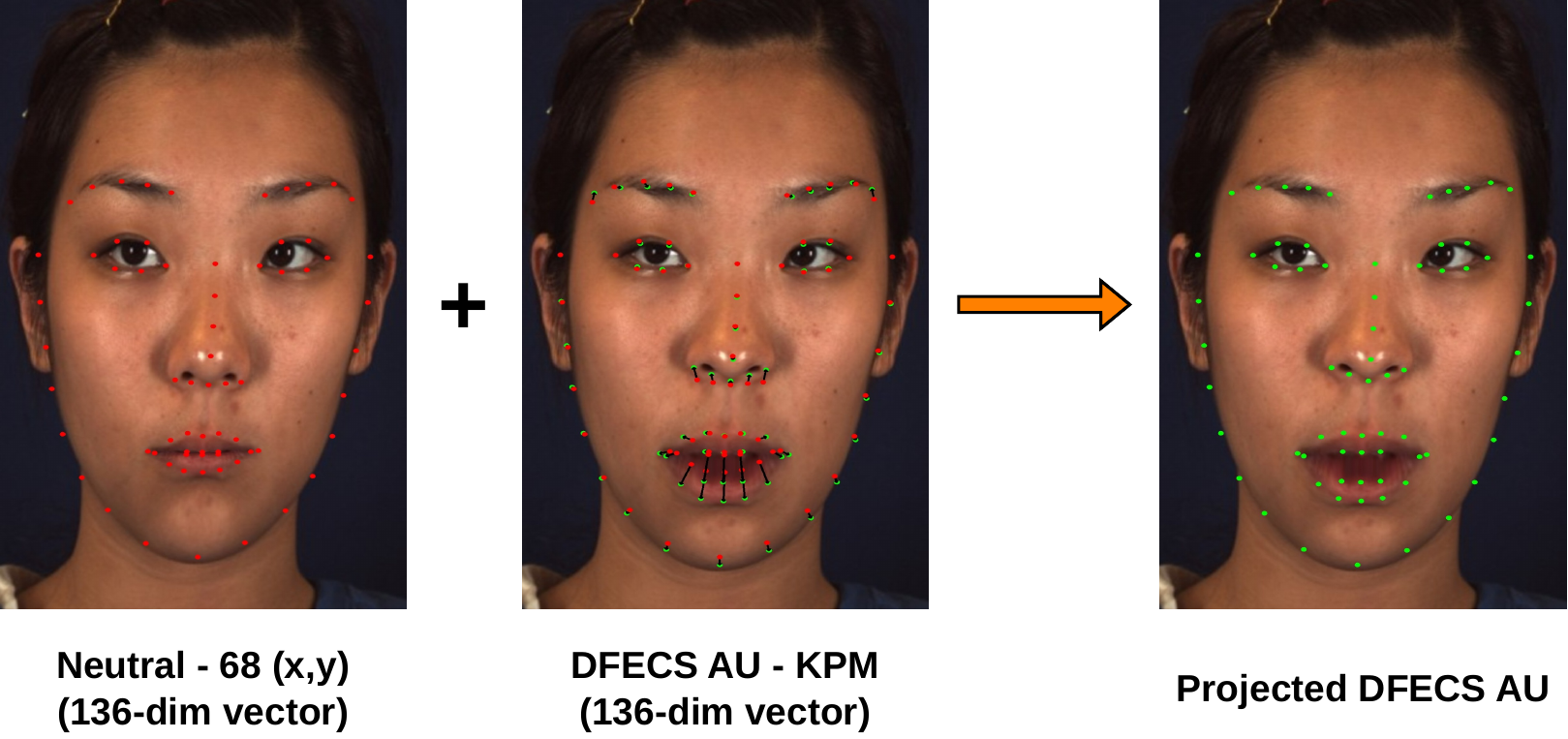}
    \caption{Generation of a projected AU image on a face. Red keypoints indicate the neutral image's facial keypoints, green ones represent the expression frame keypoints, and arrows illustrate how the keypoints move from the neutral position to the corresponding expression image. (Face image source: BP4D-Spontaneous~\cite{zhang2013high,zhang2014bp4d})}
    \label{fig:projected_AU}
\end{figure}

\subsection{Comparing AUs with increasing L1 norm}

Comparing different Action Units (FACS AUs, PCA AUs, and DFECS AUs) for their performance, i.e., variance explained on the test datasets CK+ and BP4D-Spontaneous with increasing L1 norm from L1=0 to L1=1000.

\begin{figure}[h]
    \centering
    \includegraphics[scale=0.16]{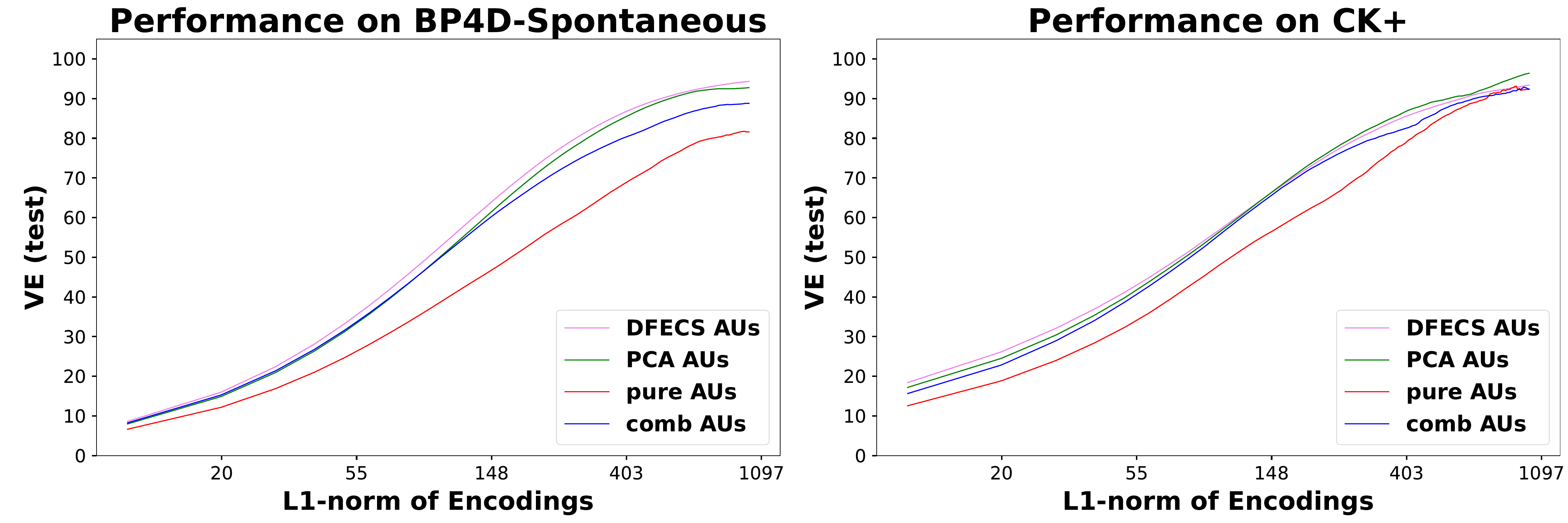}
    \caption{Performance of PCA AUs, DFECS AUs, pure AUs, and comb AUs on the test datasets - CK+ and BP4D-Spontaneous with increasing L1-norm of the encoding matrix. (x-axis on the log scale)}
    \label{fig:all_aus_L1_compare}
\end{figure}

\pagebreak
\subsection{Final NMF and DL hyperparameters for DFECS AUs}
\label{sec:hyperparameters_DFECS_AUs}

Hyperparameters for the final FFM model trained on the DISFA dataset.

\begin{table}[h]\centering
\caption{Hyperparameters for final PFM models}\label{tab:hyperparameters_PFM}
\scriptsize
\begin{tabular}{llccr}\toprule
\multicolumn{4}{c}{\textbf{DL Parameters}} \\\midrule
\textbf{Dataset} &\textbf{Face\_Part} &\textbf{Components($k_f$)} &\textbf{Sparsity ($\alpha$)} \\\midrule
DISFA (train) &left eyebrow &5 &1.15 \\
DISFA (train) &right eyebrow &5 &1.15 \\
DISFA (train) &left eye &5 &0.5 \\
DISFA (train) &right eye &5 &0.5 \\
DISFA (train) &jawline &5 &2 \\
DISFA (train) &nose &5 &1 \\
DISFA (train) &lips &5 &1.5 \\
BP4D (train) &left eyebrow &5 &2.5 \\
BP4D (train) &right eyebrow &5 &2.5 \\
BP4D (train) &left eye &4 &0.5 \\
BP4D (train) &right eye &3 &0.5 \\
BP4D (train) &jawline &NA &NA \\
BP4D (train) &nose &5 &2 \\
BP4D (train) &lips &5 &3 \\
CK+ &left eyebrow &5 &2 \\
CK+ &right eyebrow &5 &2 \\
CK+ &left eye &4 &0.5 \\
CK+ &right eye &4 &1 \\
CK+ &jawline &4 &0.5 \\
CK+ &nose &5 &2 \\
CK+ &lips &3 &2.1 \\
\bottomrule
\end{tabular}
\end{table}

\begin{table}[h]\centering
\caption{Hyperparameters for final HFM model}\label{tab:hyperparameters_HFM}
\scriptsize
\begin{tabular}{lcccc}\toprule
\multicolumn{4}{c}{\textbf{NMF Parameters}} \\\midrule
\textbf{Dataset} &\textbf{Components($q$)} &\textbf{Sparsity ($\alpha_{A}$)} &\textbf{Sparsity ($\alpha_{B}$)} \\\midrule
DISFA (train) &16 &0.1 &0.1 \\
BP4D (train) &10 &0.1 &0.1 \\
CK+ &10 &0.1 &0.1 \\
\bottomrule
\end{tabular}
\end{table}

\pagebreak

\subsection{Artifacts in projected AUs} 

\begin{figure}[h]
    \centering
\includegraphics[scale=0.20]{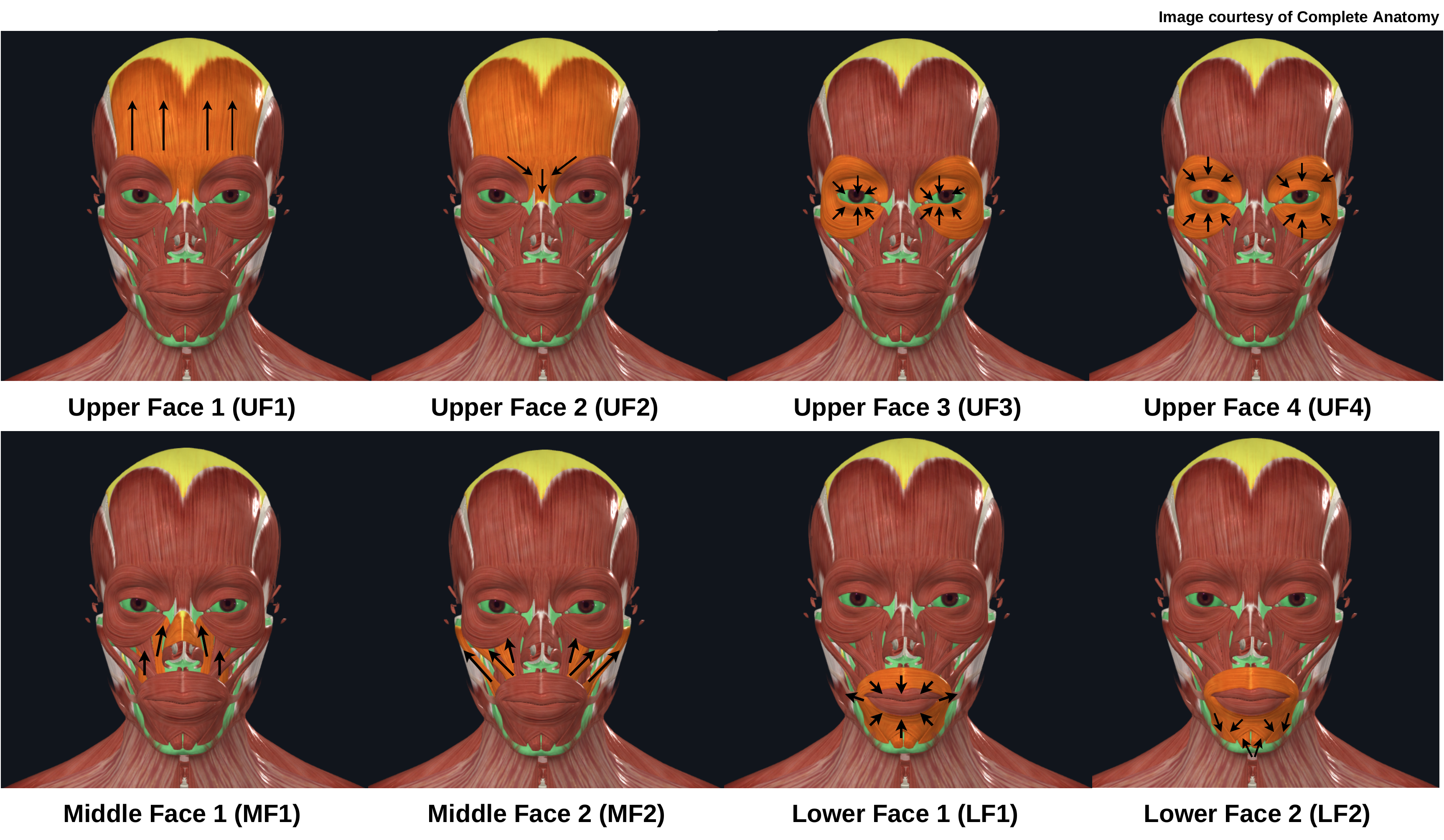}
    \caption{Muscle movements of AUs in Facial Action Coding System (FACS) excluding AUs coding for head movements.}
    \label{musclefig2}
\end{figure}

Description of the artifacts labeled by volunteer 1 in the projected AU images of PCA AUs and DFECS AUs by comparing the keypoint movements with the direction of facial muscle movements in SI Fig.~\ref{musclefig2}. Blue dots represent non-interpretable keypoint movements called artifacts in any projected AU image (see more about projected AUs in SI section~\ref{subsec:projected_AU}).

\begin{longtable}{|c|p{4cm}|p{4cm}|}
\caption{Description of artifacts in PCA components. (Face image source: BP4D-Spontaneous~\cite{zhang2013high,zhang2014bp4d})}\label{tab:desc_artifacts_pca}\\\toprule
\textbf{Component} & \textbf{Artifact}&\textbf{Muscular Basis} \\

\hline 
\endfirsthead

\hline
\multicolumn{3}{|c|}%
{{\bfseries \tablename\ \thetable{} -- continued from previous page}} \\
\hline \textbf{Component} & \textbf{Artifact}&\textbf{Muscular Basis} \\ 
\hline 
\endhead

\hline \multicolumn{3}{|r|}{{Continued on next page}} \\ \hline
\endfoot

\hline \hline
\endlastfoot

\multicolumn{3}{|c|}{\textbf{PCA - Component 2 (-ve)}} \\\midrule
\raisebox{-\totalheight}{\includegraphics[scale=0.08]{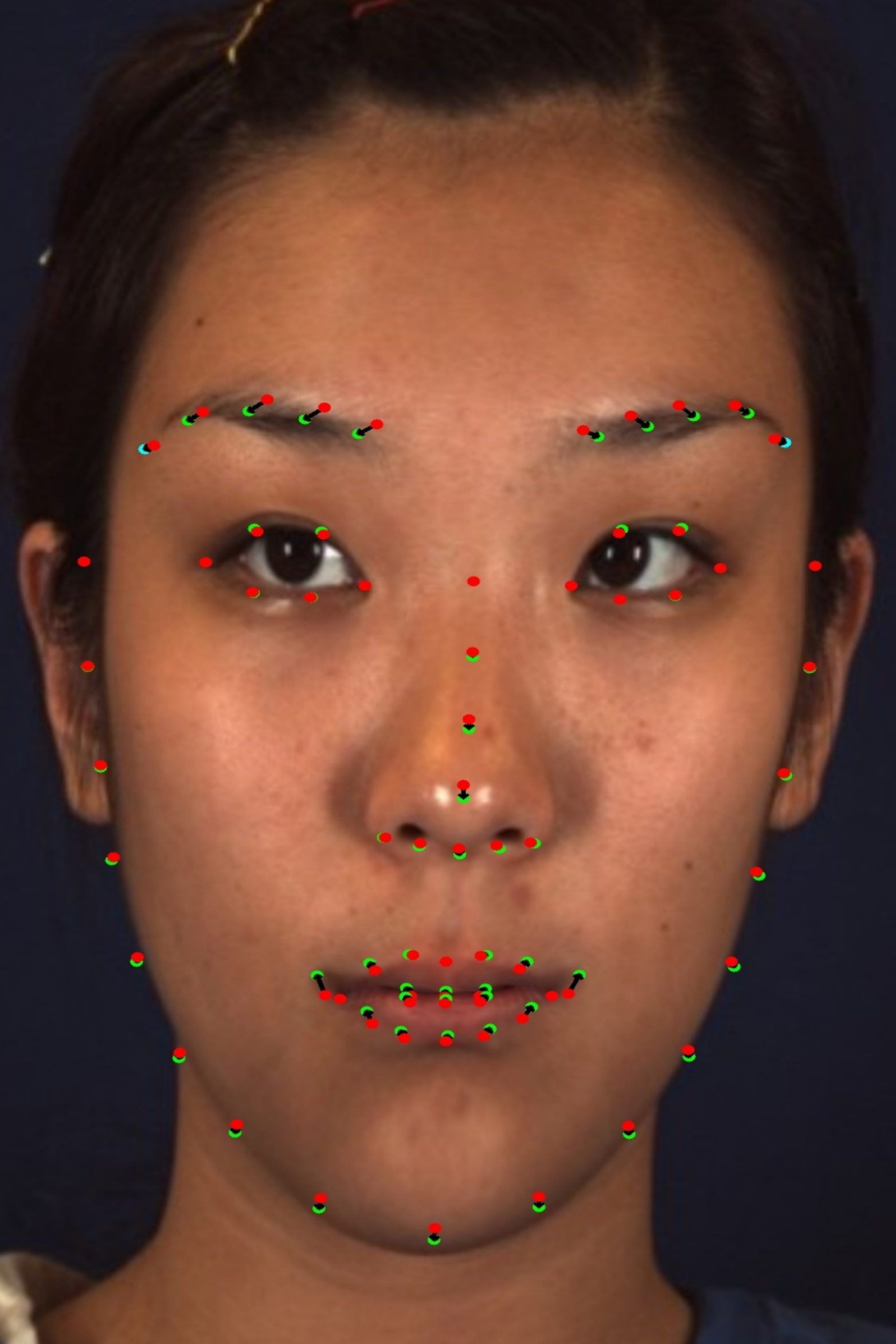}} & Eyebrows outermost point pulled diagonally towards outside & Eyebrows are either pulled in upwards (UF1), diagonally towards nose (UF2), towards eye socket (UF3,UF4) but no muscle pulling diagonally down towards outside.
\\\midrule 
\multicolumn{3}{|c|}{\textbf{PCA - Component 3 (+ve)}} \\\midrule
\raisebox{-\totalheight}{\includegraphics[scale=0.08]{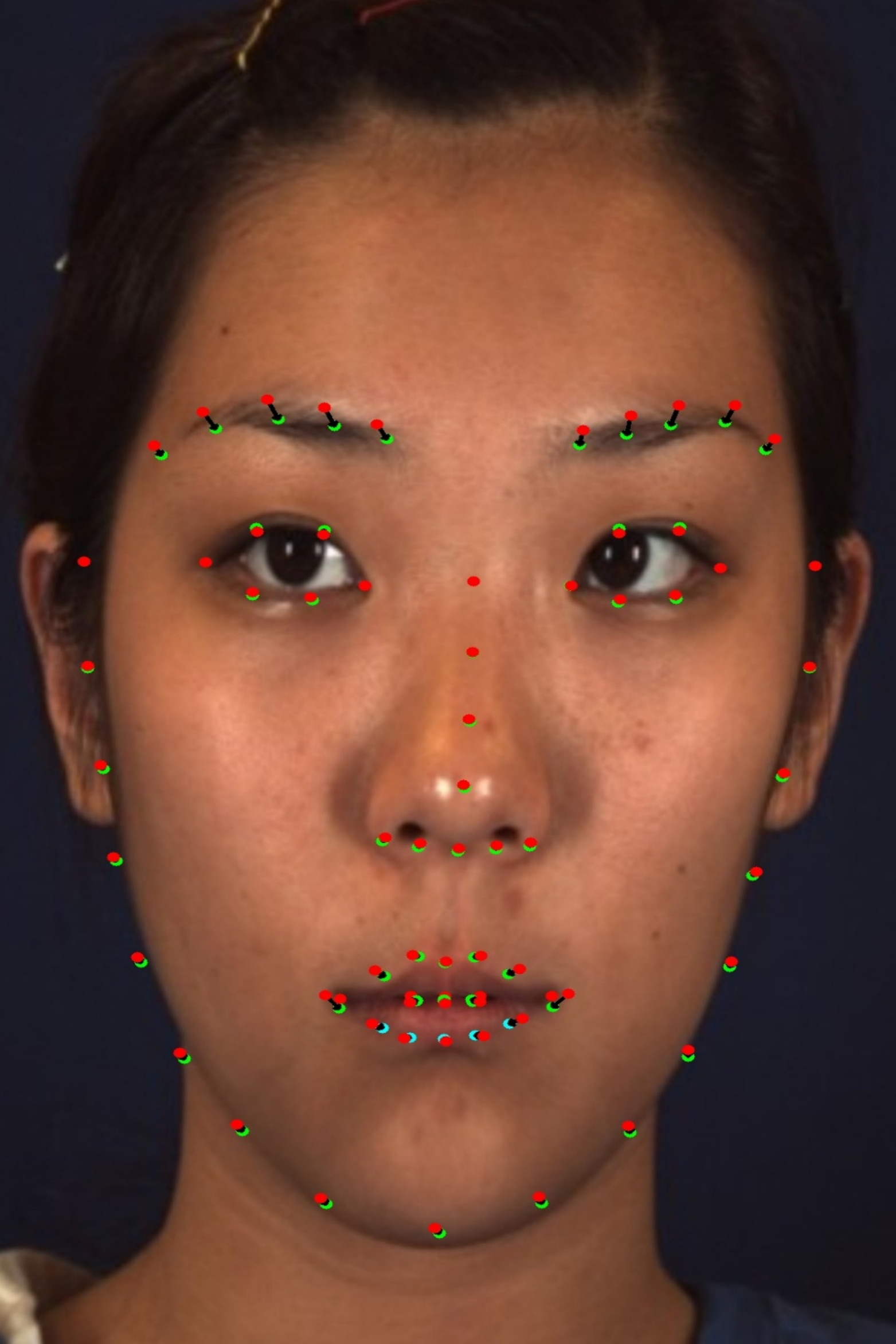}} & Lower lip keypoints moving along the line of the lower lip towards the center & Lower Lip pulled radially (LF1), sideways (LF1) or downwards (LF2) but not along the line of lower lip.
\\\midrule 
\multicolumn{3}{|c|}{\textbf{PCA - Component 5 (-ve)}} \\\midrule
\raisebox{-\totalheight}{\includegraphics[scale=0.08]{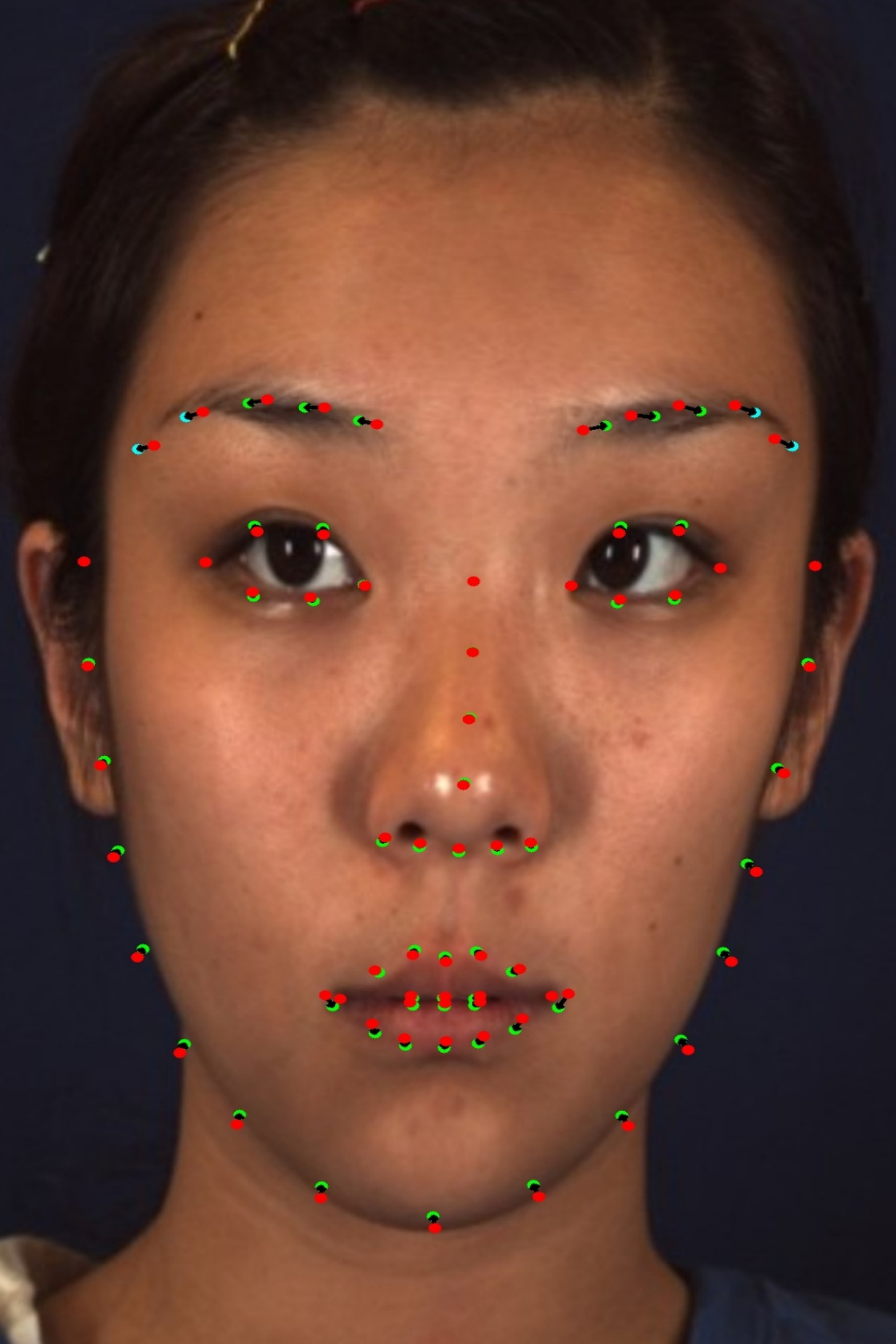}} & Eyebrow keypoints pulled almost horizontally outwards & Eyebrows are either pulled in upwards (UF1), diagonally towards nose (UF2), towards eye socket (UF3,UF4) but no muscle pulling horizontally towards outside.
\\\midrule 
\multicolumn{3}{|c|}{\textbf{PCA - Component 6 (+ve)}} \\\midrule
\raisebox{-\totalheight}{\includegraphics[scale=0.08]{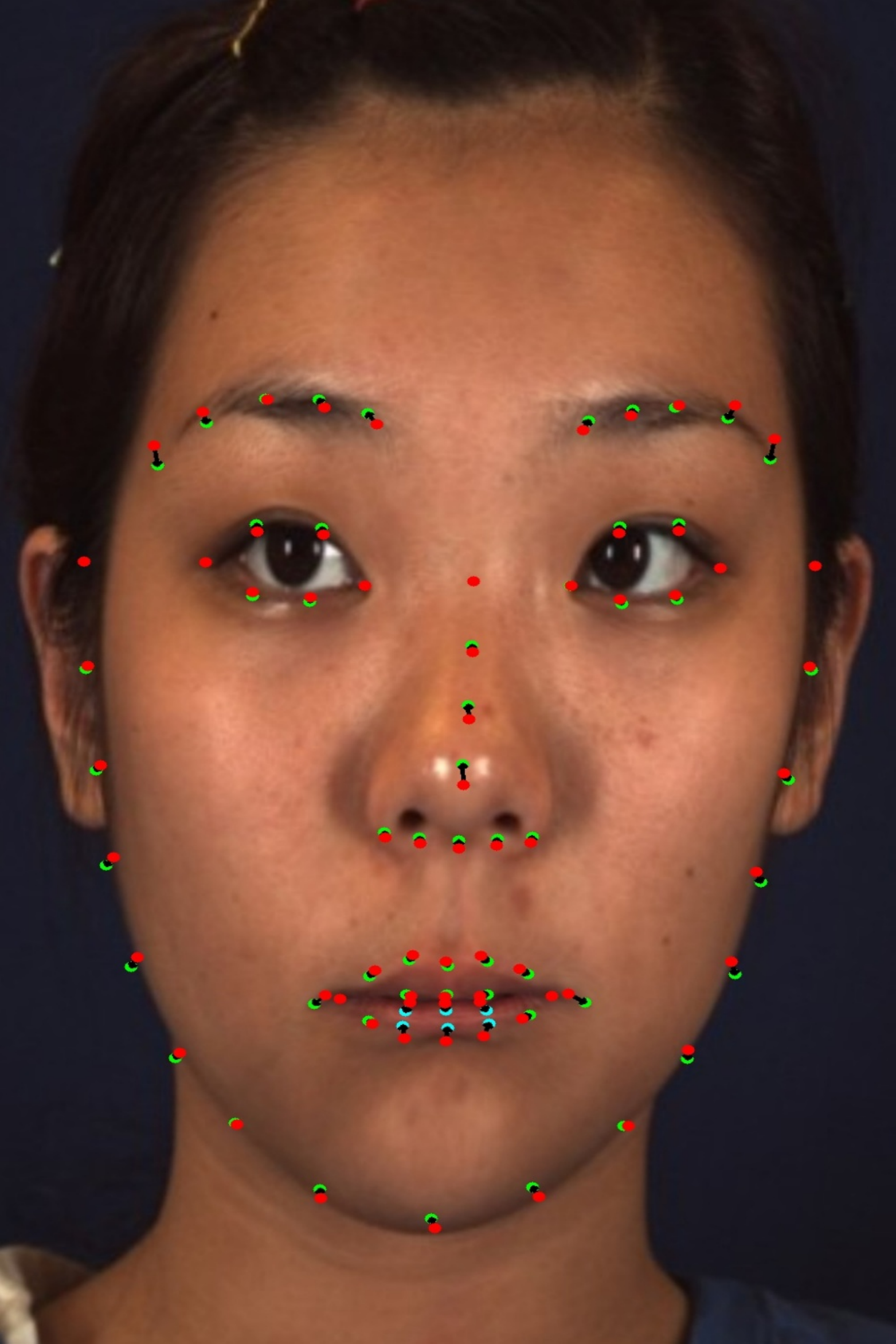}} & Lower Lip lower and upper keypoints compressing towards each other &No muscles in LF1,LF2 compresses the upper part of lower lip towards the lower part\\\midrule
\pagebreak
\multicolumn{3}{|c|}{\textbf{PCA - Component 6 (-ve)}} \\\midrule
\raisebox{-\totalheight}{\includegraphics[scale=0.08]{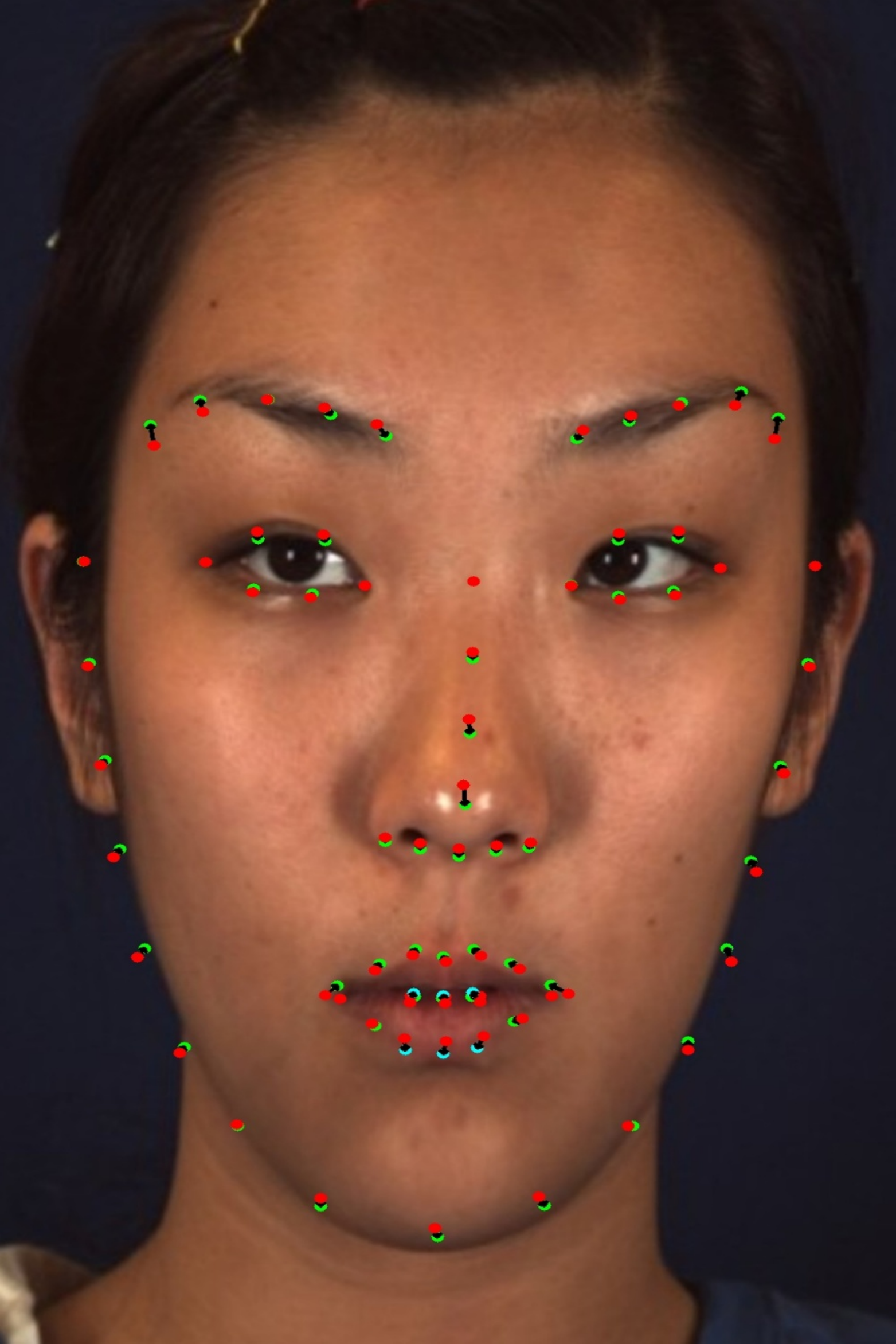}} & Lower lip upper and lower part moving in opposite direction expanding the lower lip & Lower Lip pulled radially (LF1), sideways (LF1) or downwards (LF2) but no muscle pulls the lower lip upper and lower part against each other
\\\midrule 
\multicolumn{3}{|c|}{\textbf{PCA - Component 7 (+ve)}} \\\midrule
\raisebox{-\totalheight}{\includegraphics[scale=0.08]{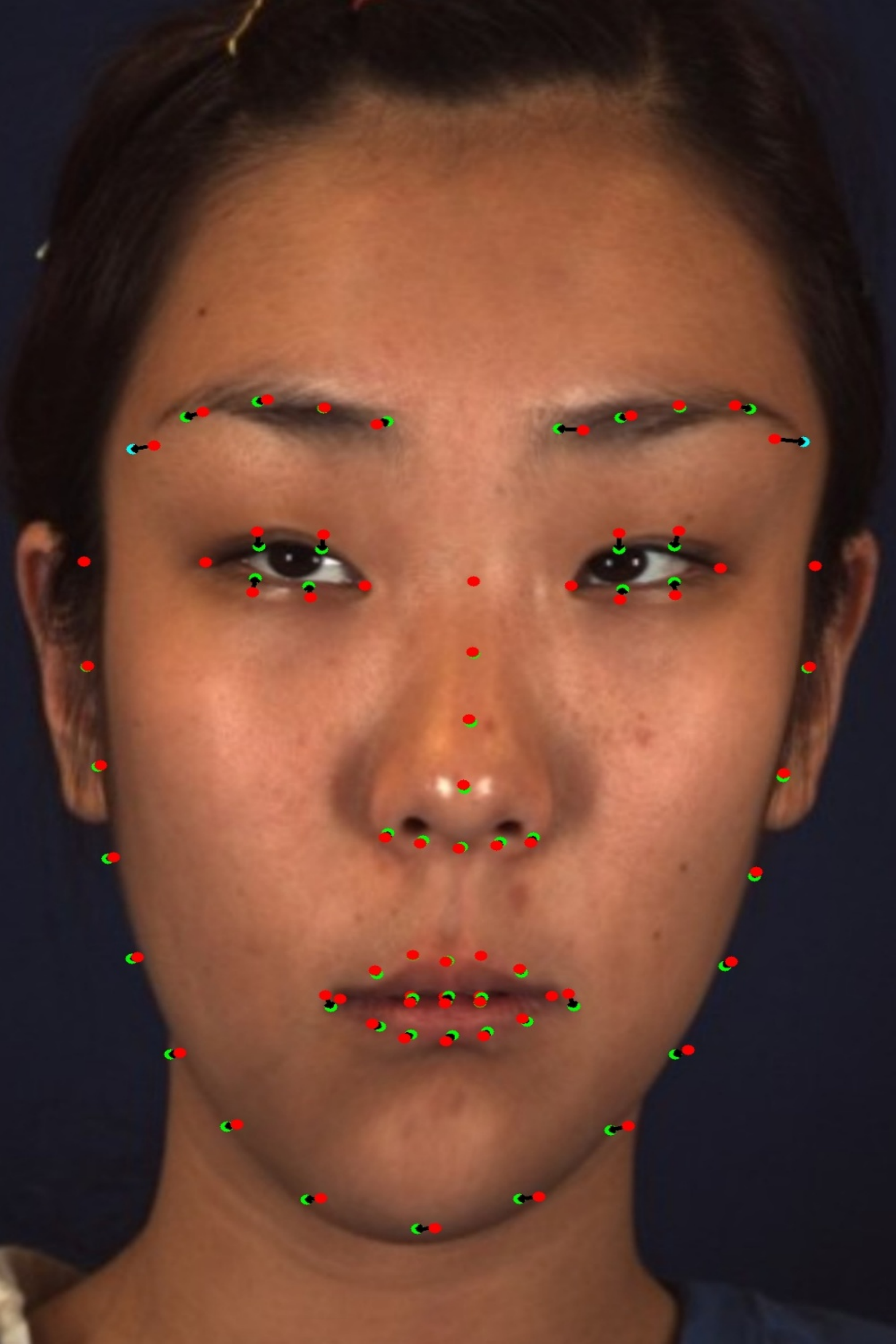}} & Eyebrows outermost keypoint pulled in a horizontal fashion towards outside &Eyebrows are either pulled in upwards (UF1), diagonally towards nose (UF2), towards eye socket (UF3,UF4) but no muscle pulling horizontally towards outside\\\midrule
\multicolumn{3}{|c|}{\textbf{PCA - Component 8 (+ve)}} \\\midrule
\raisebox{-\totalheight}{\includegraphics[scale=0.08]{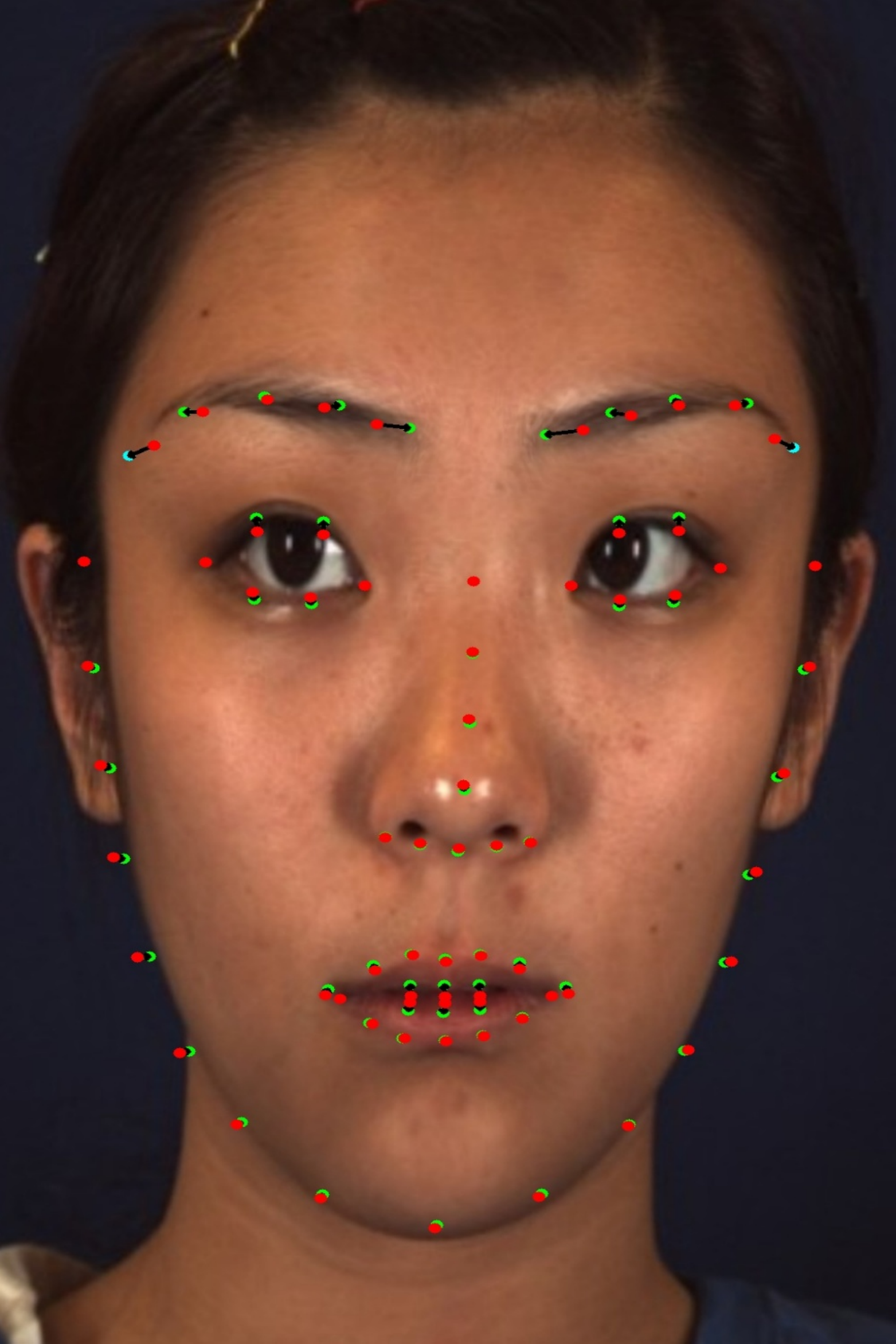}} & Eyebrows outermost point pulled diagonally towards outside & Eyebrows are either pulled in upwards (UF1), diagonally towards nose (UF2), towards eye socket (UF3,UF4) but no muscle pulling diagonally down towards outside.\\
\bottomrule
\end{longtable}

\begin{table}[h]\centering
\caption{Description of artifacts in FFM components. (Face image source: BP4D-Spontaneous~\cite{zhang2013high,zhang2014bp4d})}\label{tab:desc_artifacts_ffm}
\begin{tabular}
{|p{3cm}|p{4cm}|p{4cm}|}\toprule
\textbf{Component} & \textbf{Artifact}&\textbf{Muscular Basis} \\\midrule
\multicolumn{3}{|c|}{\textbf{FFM - Component 6}} \\\midrule
\raisebox{-\totalheight}{\includegraphics[scale=0.10]{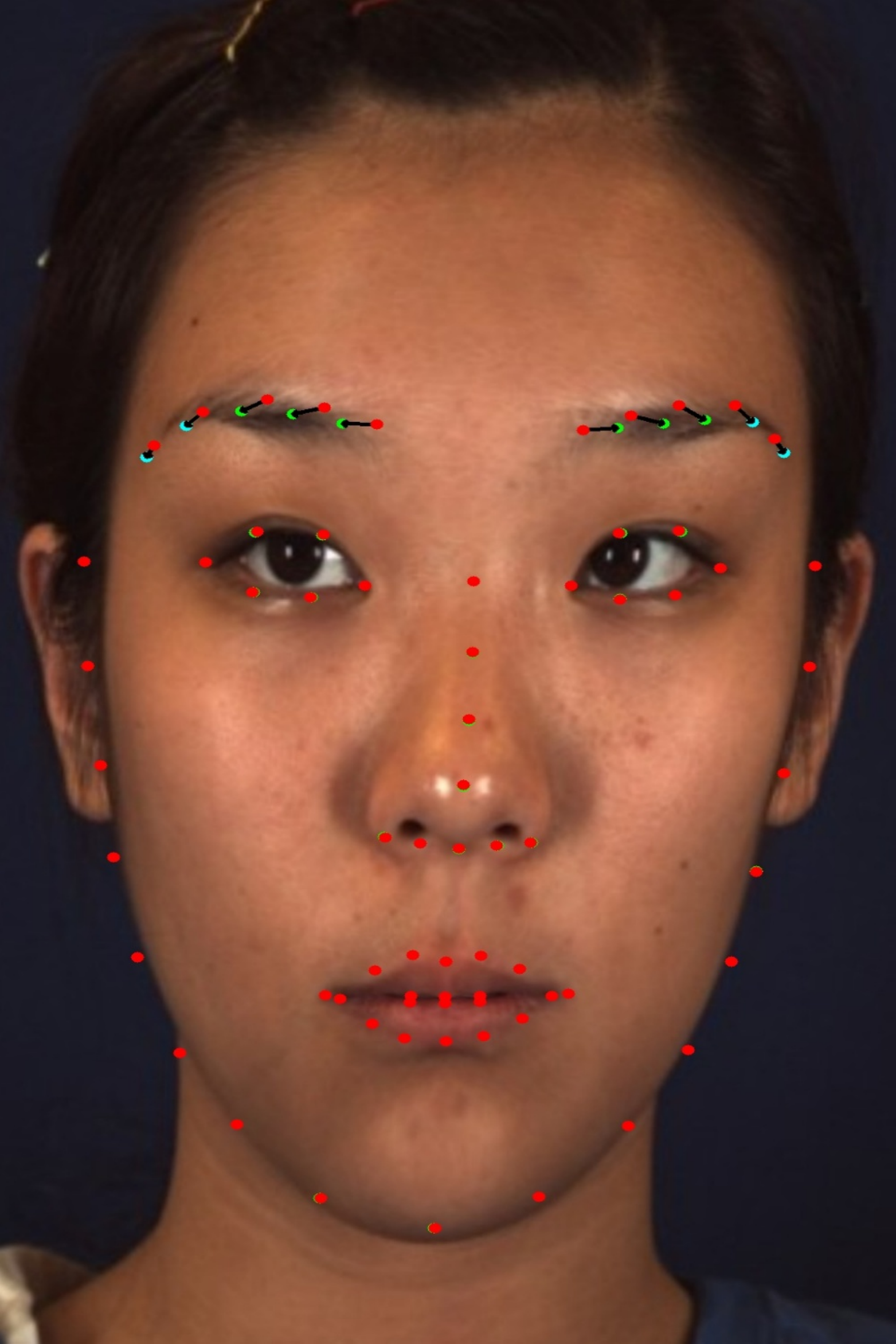}} & Eyebrow keypoints pulled almost horizontally outwards & Eyebrows are either pulled in upwards (UF1), diagonally towards nose (UF2), towards eye socket (UF3,UF4) but no muscle pulling horizontally towards outside \\\midrule 
\multicolumn{3}{|c|}{\textbf{FFM - Component 9}} \\\midrule 
\raisebox{-\totalheight}{\includegraphics[scale=0.10]{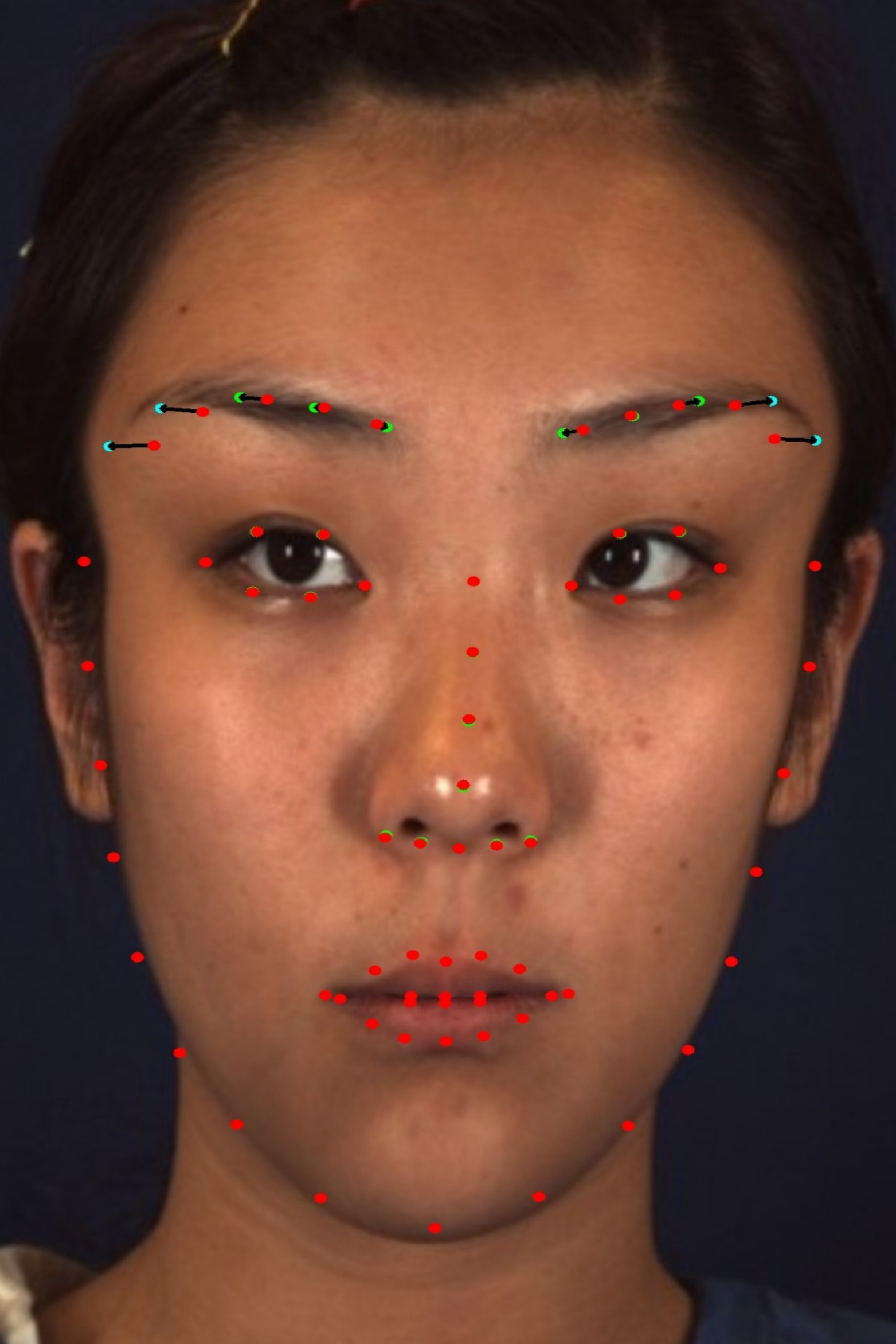}} & outermost eyebrow moving outwards in horizontal fashion & Eyebrows are either pulled in upwards (UF1), diagonally towards nose (UF2), towards eye socket (UF3,UF4) but no muscle pulling horizontally towards outside \\\midrule 
\end{tabular}
\end{table}


\end{document}